\documentclass{article}
\usepackage{ijcai22}

\usepackage{times}

\usepackage{soul}
\usepackage{url}
\usepackage[hidelinks]{hyperref}
\usepackage[utf8]{inputenc}
\usepackage[small]{caption}
\usepackage{graphicx}
\usepackage{amsmath}
\usepackage{booktabs}
\usepackage{enumitem}
\usepackage[ruled,vlined, linesnumbered]{algorithm2e}

\usepackage{amssymb,amsthm}
\usepackage{amsfonts}       
\usepackage{multirow}
\usepackage{bm}
\usepackage{epsfig}
\usepackage{epstopdf}
\usepackage{lipsum}
\usepackage{mathtools}
\usepackage{microtype}      
\usepackage{nicefrac}       
\usepackage[labelformat=simple]{subcaption}
\usepackage{xcolor}         

\usepackage{float}

\newcommand{\independent}{\!\perp\!\!\!\perp}

\newtheorem{theorem}{Theorem}

\newtheorem{corollary}{Corollary}

\newlist{myitemize}{itemize}{3}
\setlist[myitemize,1]{label=$\circ$,leftmargin=1em}
\setlist[myitemize,2]{label=$\circ$,leftmargin=2em}
\setlist[myitemize,3]{label=$\circ$}

\title{NeuCEPT: Locally Discover Neural Networks' Mechanism \\ via Critical Neurons Identification with Precision Guarantee}

\author{
	Minh N. Vu
	\and
	Truc D. Nguyen
	\And
	My T. Thai
	\affiliations
	{University of Florida},
	Gainesville, Florida, USA
	\emails
	minhvu@ufl.edu, truc.nguyen@ufl.edu, mythai@cise.ufl.edu
}

\begin{document}

\maketitle

\begin{abstract}
Despite recent studies on understanding deep neural networks (DNNs), there exists numerous questions on how DNNs generate their predictions. Especially, given similar predictions on different input samples, are the underlying mechanisms generating those predictions the same? 
In this work, we propose NeuCEPT, a method to locally discover critical neurons that play a major role in the model's predictions and identify model's mechanisms in generating those predictions. We first formulate a critical neurons identification problem as maximizing a sequence of mutual-information objectives and provide a theoretical framework to efficiently solve for critical neurons while keeping the precision under control. NeuCEPT next heuristically learns different model's mechanisms in an unsupervised manner. 
Our experimental results show that neurons identified by NeuCEPT not only have strong influence on the model's predictions but also hold meaningful information about model's mechanisms. 
\end{abstract}

\section{Introduction}

\begin{figure}[ht]
     \centering
         \includegraphics[width=0.7\linewidth]{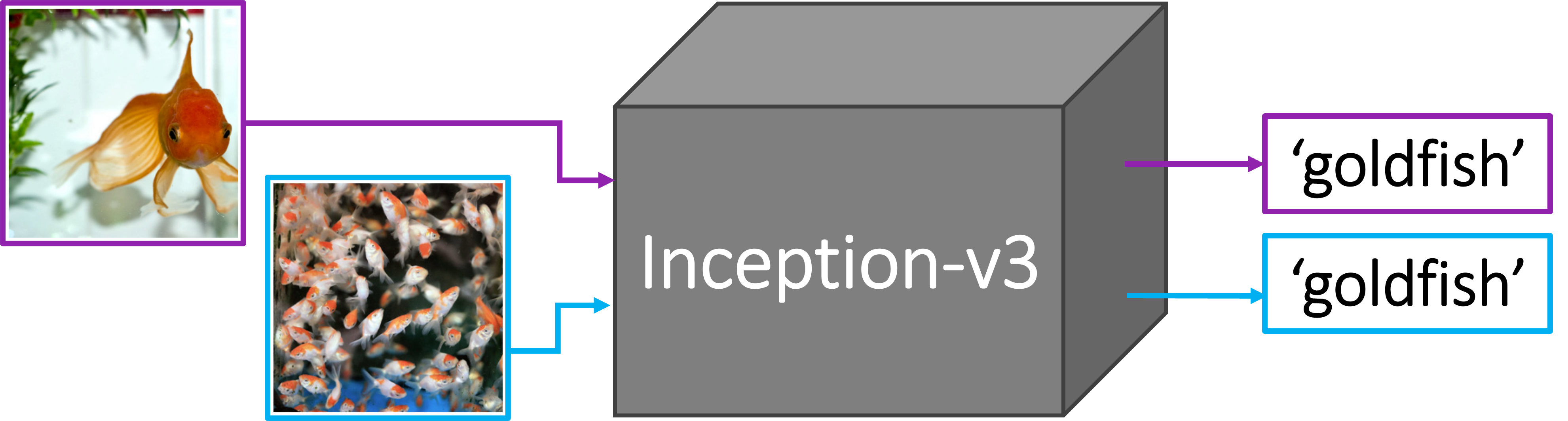}
     \caption{Two images have the same predictions \textit{goldfish} generated by Inception-v3. One is a \textit{single goldfish} while the other is a \textit{shoal of fish}. Are the mechanisms behind the two predictions the same?}
     \label{fig:problem}
 \end{figure}

 Significant efforts have been dedicated to improve the interpretability of modern neural networks, leading to several advancements~\cite{Lipton2016IML,Murdoch22071}; however, few works have been conducted to characterize and analyze local prediction of the neural networks based on the internal forwarding process of the model (see Related Work).
 In this paper, we focus on investigating different mechanisms learnt by the neural networks to generate predictions (see Fig.~\ref{fig:problem} as an example). Intuitively, the mechanism of a prediction is the forwarding process producing the prediction in the examined model (see concrete definition in Sect. \ref{sec:learning}). Our hypothesis is that predictions of the same class label can be generated by different mechanisms which can be captured and characterized by activation of some specific neurons, called \textit{critical neurons}. Analyzing the activation of those neurons can help identify the model's mechanisms and shed light on how the model works. 
 
\begin{figure}[ht]
     	\centering
			\includegraphics[width=0.8\linewidth]{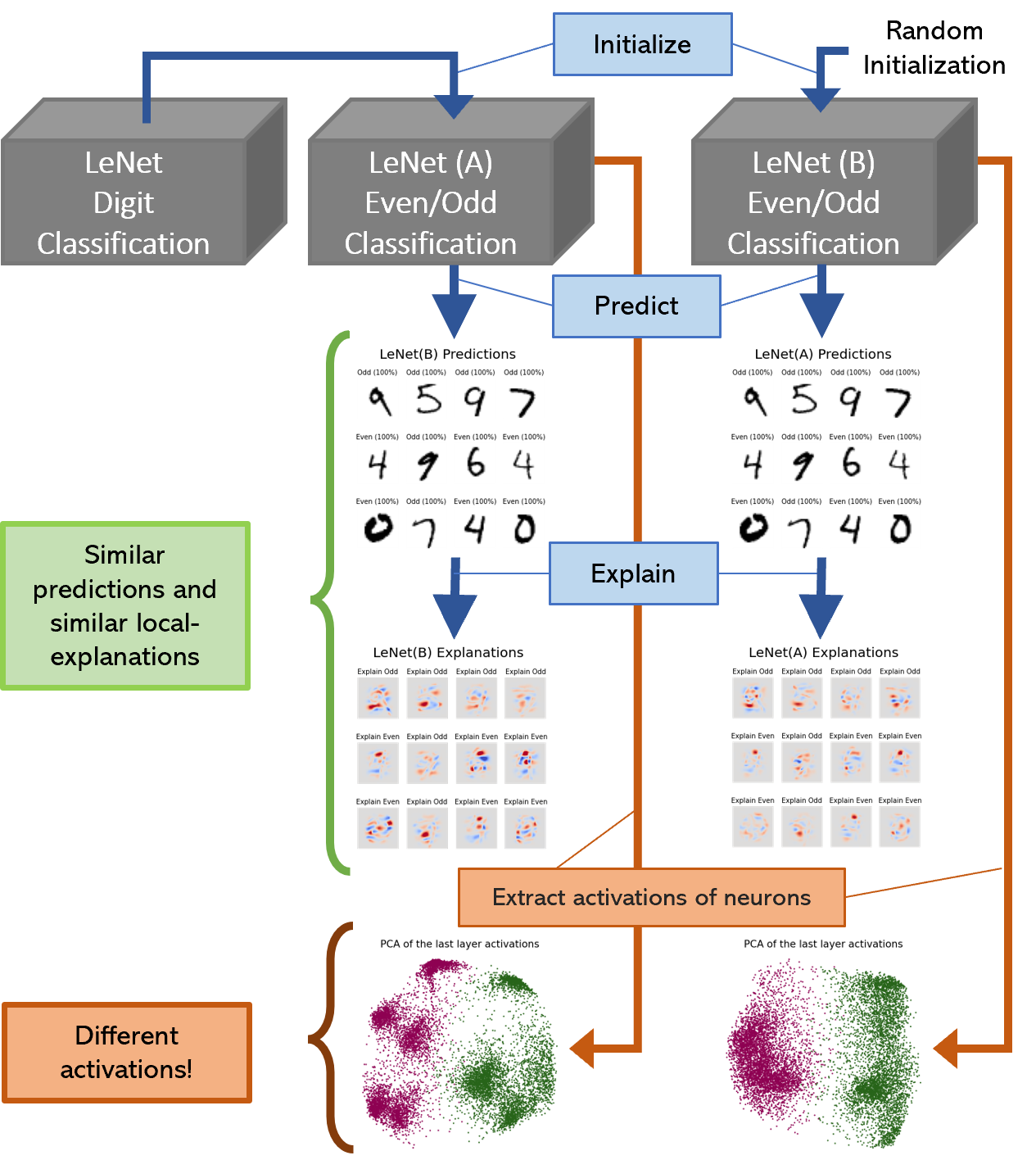}
		\caption{{LeNet(A) is initialized by a LeNet pretrained on digit classification task while LeNet(B) is initialized randomly. While the \textit{even/odd} predictions and the explanations provide little information differentiating the two models, extracting and visualizing the activation at the models' last layer reveal that LeNet(A) groups inputs into more distinctive clusters with the same ground-truth digit-labels.}}
		\label{fig:dnn}
 \end{figure}

 Following are some key reasons motivating our study:
  
  \begin{itemize}
    \item The identification and study of critical neurons can serve as an initial model’s examination, which reveals new aspects on the model's dynamics. For instance, Fig.~\ref{fig:dnn} shows two LeNet trained on the MNIST digit dataset to classify \textit{even} or \textit{odd} digit. While the models' outputs and feature-based local explanations do not suggest significant differences between the two models, the evidences based on critical neurons
    clearly illustrate that predictions of the two models are generated differently. Appendix~\ref{appendix:linearprobe} provides further discussion on how critical neurons' analysis can strengthen the examination of the layers' linear separability~\cite{alain2017understanding}.
    \item Critical neurons allow us to characterize model’s predictions based on how they are generated by the model. Each set of similar predictions can be studied and analyzed for downstream tasks such as performance~\cite{Everingham2009ThePV} and trust evaluation~\cite{Wang2020ASO}. Appendix~\ref{append:chexnet} provides an example of our analysis on CheXNet~\cite{CheXNet}, a modern DNN which can detect pneumonia from chest X-rays at a level exceeding practicing radiologists. It shows how understanding the mechanism can help identify unreliable predictions.
    \item Compared to local attribution explanation methods, the study of critical neurons provide a new dimension on how we explain the predictions. Specifically, as demonstrated in Appendix~\ref{append:inception}, NeuCEPT, our algorithm based on critical neurons, identifies certain neurons holding vital information in characterizing model's predictions of given class labels. Importantly, those neurons are not among the top-activated neurons of each class and they are not normally identified by attribution methods due to their low average activation level.
\end{itemize}

\begin{figure*}[ht]
		\centering
			\includegraphics[width=0.9\linewidth]{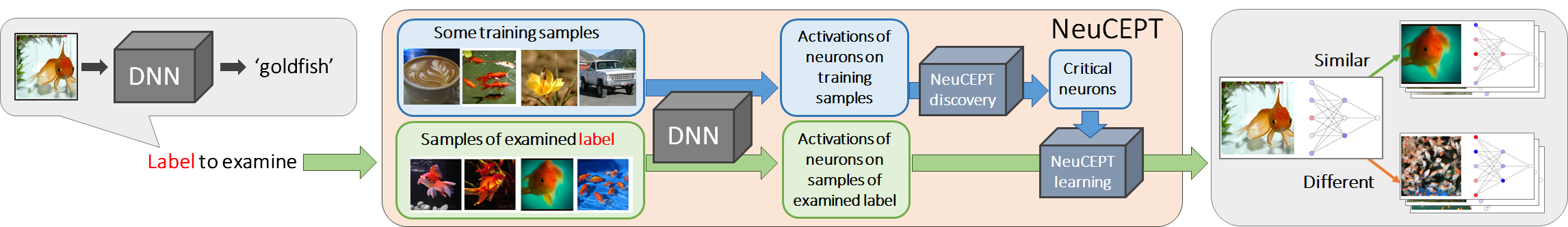}
		\caption{{Overall architecture of NeuCEPT. Given a prediction of a class of interest, NeuCEPT collects the model's inputs from that class. Then, by forwarding them through the DNN, NeuCEPT obtains the activation and solves for the set of critical neurons at some layers. Finally, mechanisms are learnt via unsupervised learning on those neurons' activation.}}
		\label{fig:architecture}
\end{figure*}

 We propose NeuCEPT - a method to locally discover \underline{neu}ral network's mechanism via \underline{c}ritical n\underline{e}urons identification with \underline{p}recision guaran\underline{t}ee, as shown in Fig.~\ref{fig:architecture}. The innovation of NeuCEPT lies in its two main components: NeuCEPT-discovery and NeuCEPT-learning, with the following contributions:  
\begin{myitemize}
    \item {\bf Critical Neurons Identification with NeuCEPT-discovery (Sect. \ref{section:neuronselection}).} Under an information-theoretic point of view, we introduce a layer-by-layer mutual information objective to discover the \textit{critical neurons} of model's predictions. Intuitively, under this objective, the critical neurons of a layer is the set of neurons that determine the state of the critical neurons of the sub-sequence layer, which eventually determine the model's output. To overcome the high running-time complexity in solving for this objective, we further develop a theoretical framework to approximate a set of critical neurons in a pairwise-parallel manner while keeping the precision guarantee.
    \item {\bf  Information-theoretic interpretation and NeuCEPT-learning of Critical Neurons (Sect. \ref{sec:learning}).} We provide Information-theoretic interpretation of the critical neurons and elaborate how learning the mechanism on top of critical neurons can result in better claims on DNNs' mechanisms. We propose an unsupervised learning procedure, called NeuCEPT-learning, to carry out that task. 
    \item {\bf Prior-knowledge Training and Experimental Results (Sect. \ref{sec:experiments}).} We propose a new testing approach, the prior-knowledge training, to experimentally evaluate the claims on DNNs' predicting mechanism. In this training, mechanisms are embedded in the prior-knowledge trained (PKT) model. Our rigorous experimental results on MNIST and CIFAR-10 show that NeuCEPT is consistently able to detect the embedded mechanisms in PKT models. Ablation study is also conducted extensively.
\end{myitemize}

\textbf{Related Work.} \label{sec:related}
 Although the problem of identifying DNN's mechanism is not well-studied, the related researches of interpretability and neural debugging received a lot of attention in recent years. Regarding explanation methods, there are two major approaches: local and global~\cite{Lipton2016IML}. 
 
 Local methods either search for an interpretable surrogate model to capture the local behavior of the DNNs~\cite{Scott2017,Marco2016,minhpgm} or carefully back-propagate the model~\cite{Avanti2017,Simonyan2013,Mukund2017} to attribute contribution scores. The focus of these methods is on identifying features and, technically can be extended to, neurons with highly-attributed scores; however, it is unclear how they can be exploited to identify model's mechanism. Specifically, high attributed model's neurons do not necessarily imply the capability to identify mechanism\footnote{Examples in Appendix~\ref{append:inception} 
 show how highly activated features fail to differentiate predictions with different mechanisms.}.
 
 On the other hand, global methods focus on the model's behavior globally, using decision trees~\cite{DistillingToTree}, decision sets~\cite{lakkaraju2016interpretable} or recursive-partitioning~\cite{Yang2018}. Unfortunately, there has been no well-established connection between the explained global dynamics and the model's local prediction; hence, the question regarding local prediction's mechanism is still unanswered. Among global methods, we find~\cite{Wang2018} to be the most related research to our work. Specifically, the method uses distillation technique to train another model with sparser data forward paths and associates each input with one. Even though the paths can be used to partially reveal the model's mechanism, they are extracted from the distilled model and there is no guarantee that they maintain the mechanism behind the predictions of the original model. Furthermore, the method requires to retrain the model and also model-specific. NeuCEPT, on the other hand, is applicable as long as the class data is sufficient and the model's activation can be extracted. 

\section{Critical Neurons Identification with NeuCEPT-discovery} \label{section:neuronselection}

Although modern DNNs contain from thousands to millions neurons, 
there is evidence that only a small portion of neurons contributes mostly to the predictions~\cite{Bau2020}. We denote such neurons as \textit{critical neurons of the predictions} or \textit{critical neurons} for short. Identifying critical neurons not only reduces the analysis's complexity but also offers more compact explanations for model's mechanism on making its predictions. Unfortunately, due to the sequential structure of DNNs, identifying such critical neurons is a daunting task, from formulating a suitable objective function to solving the problem and interpreting those neurons' activation.

\subsection{Problem Formulation}\label{subsec:formulation}
We consider DNNs in classification problems with the forwarding function $y = f(x)$, where $ y \in \mathbb{R}^m$ is a logit and $ x \in \mathbb{R}^n$ is an input. The neural network has a sequential structure with $L$ layers, each layer has $k_l$ neurons $(l=1,...,L)$. The activation of neurons at any layers on a given input can be computed by forwarding the model. We denote this computation as $z_l = f_l(x)$ 
where $f_l: \mathbb{R}^n \rightarrow \mathbb{R}^{k_l}$. Then $z = [z_0, ..., z_L]$ is the activation of all model's neurons on that input.
We use capital letters to refer to random variables, i.e. $Z_l$ is the random variable representing the activation of neurons at layer $l$. The superscript notation refers to the random variables associated with a subset of neurons. For instance, given a subset $\mathcal{S}$ of neurons at layer $l$, $Z_l^{\mathcal{S}}$ is the random activation of the neurons in $\mathcal{S}$ at layer $l$. Due to the forwarding structure of the model, the activation of neurons at a given layer depends only on the activation of neurons at the previous layer, i.e. $Z_l \independent Z_j | Z_{l-1}, \forall j = 0,...,l-2$, where $\independent$ denotes the independent relationship. Thus, we have the Markov chain:
\begin{align}
    X = Z_0 \rightarrow Z_1 \rightarrow \cdots \rightarrow Z_L.
    \label{eq:markov}
\end{align}
Given a prediction, a layer $l$ and the corresponding random activation $Z_l$, we would like to identify a subset of the critical neurons at that layer, i.e. the subset of neurons containing the most information on the prediction of interest.
From a information-theoretic approach, we formalize the notion of criticality using mutual-information (MI) and formulate the critical neurons identification (CNI) problem as:
\begin{align}
        \mathcal{S}_l = \textup{argmax}_{\mathcal{S} \subseteq \mathcal{N}_l} I\left(Z_l^{\mathcal{S}};Z_{l+1}^{\mathcal{S}_{l+1}} \right), \textup{ s.t. } \mathcal{S} \in \mathcal{C} \label{eq:trueobjective}
\end{align}
where $\mathcal{N}_l$ is the set of neuron at layer $l$, $I$ is the joint mutual-information function~\cite{cover2006} and $\mathcal{S} \in \mathcal{C}$ represents some complexity constraints imposed on the problem. 
The intuition of this formulation is, at each layer, we search for the set of neurons holding the most information on the set of neurons solved in the next layer. We denote $o$ as the neuron associated with the class of prediction and $Y = Z_{L}^{\{ o \}}$ the corresponding random activation. By bounding the first optimization at the last layer $L$ to maximize $ I\left(Z_{L-1}^{\mathcal{S}};Y \right)$, we enforce the sub-sequence optimizations at the earlier layers to solve for the set of neurons holding the most information on the prediction of interest. Sect.~\ref{subsec:intuition} provides more information-theoretic intuition of this objective.

\subsection{Solutions with Precision Guarantee} \label{subsect:select}

As CNI is in NP-hard\footnote{Our proof of CNI is in NP-hard is shown in Appendix~\ref{proof:nphard}}, we introduce an algorithm NeuCEPT-discovery to approximate CNI with precision guarantee. As an abstract level, NeuCEPT-discovery considers each pair of a layer's activation-output of the examined DNN as an input-response pair and conducts critical neuron selection on them. Different from the sequential formulation of CNI (\ref{eq:trueobjective}), NeuCEPT-discovery is executed in a pair-wise manner and, consequently, can be implemented efficiently. Our theoretical results guarantee that the modification to pair-wise optimization still achieves the specified precision level.

{\bf Using Markov Blanket.} Given a random variable $T$, we denote $\mathcal{M}_l (T) \subseteq \mathcal{N}_l$ as the smallest set of neurons at layer $l$ such that, conditionally on the variables in that set - $Z_l^{\mathcal{M}_l (T)}$, $T$ is independent of all other variables at layer $l$. The set $\mathcal{M}_l (T)$ is commonly addressed as the Markov blanket (MB)\footnote{There is an ambiguity between the notions of \textit{Markov-blanket} and \textit{Markov-boundary}. We follow the notion of the \textit{Markov-blanket} defined in \cite{Koller2009}, which is consistent with~\cite{Candes2016}. In fact, the Markov-blanket of a random variable $T$ in joint distribution ${P}$, i.e. $\mathcal{M}(T)$, is the minimum set of variables such that, given realizations of all variables in $\mathcal{M}(T)$, $T$ is conditionally independent from all other variables.} of $T$ in the studies of graphical models. We make a slight modification by restricting the set of variables to a certain layer of the model. Under very mild conditions about the joint distribution of $T$ and $Z_l$, the MB is well defined and unique~\cite{PearlPaz_2014}. In this work, we follow researchers in the field, assume these conditions~\cite{edward2000} and proceed from there.

Substituting $T$ by ${\displaystyle Z_{l+1}^{\mathcal{S}_{l+1}}}$, we have ${\displaystyle \mathcal{S} = \mathcal{M}_l \left(Z_{l+1}^{\mathcal{S}_{l+1}} \right)}$, the MB at layer $l$ of $Z_{l+1}^{\mathcal{S}_{l+1}}$, achieves the maximum of the objective (\ref{eq:trueobjective}) since the MB contains all information about $Z_{l+1}^{\mathcal{S}_{l+1}}$. On the other hand, by the definition of the MB, ${\displaystyle \mathcal{M}_l \left(Z_{l+1}^{\mathcal{S}_{l+1}} \right) }$ is the smallest subset maximizing the MI. This aligns with our intuition that the MB are important to the activation of critical neurons at later layers. Using the MB, we have a straight approach to solve (\ref{eq:trueobjective}): Given the activation of interest at the last layer $Y = Z_L^{\{ o\}}$, we solve for $\mathcal{M}_{L-1} \left( Y \right)$ - the MB at layer $L-1$. Then, at layer $L-2$, we find the MB of the variables in $\mathcal{M}_{L-1} \left(Y \right)$. The process continues until the first layer is reached, which can be described as:
\begin{align}
    \mathcal{S}_{L-1} \leftarrow \mathcal{M}_{L-1} \left(Y \right), \quad \mathcal{S}_{l-1} \leftarrow  \mathcal{M}_{l-1} \left( Z_l^{\mathcal{S}_{l}} \right) \label{eq:back_propagate}
\end{align}

{\bf Controlling the precision.} Unfortunately, directly solving (\ref{eq:back_propagate}) is impractical as the problem is also in NP-hard~\cite{DimitrisNIPS1999}. Additionally, due to the curse-of-dimensionality, sampling enough data to estimate the distribution of neurons' activation $Z_l^{\mathcal{S}_{l}}$ accurately is impractical. Our key observation to overcome these challenges is that the MB of the model's output variable $Y$ at each layer $l$ is a subset of $\mathcal{S}_l$ defined in procedure (\ref{eq:back_propagate}). As a result, given a solver solving for $\mathcal{M}_{l} \left(Y \right)$ with a precision at least $p$, the output of that solver is also an approximation of $\mathcal{S}_{l}$ with the precision at least $p$. This allows us to solve for $\mathcal{M}_{l} \left(Y \right)$ instead of $\mathcal{S}_{l}$ and overcome the high-dimensionality of $Z_l^{\mathcal{S}_{l}}$. These key steps of NeuCEPT-discovery are described in Algorithm~\ref{algo:fdr_ko} (Appendix~\ref{sect:algo}).
The proof that NeuCEPT-discovery achieves precision guarantee is based on Theorem~\ref{eq:subset_cond}, which states $\mathcal{M}_{l} \left(Y \right)$ is a subset of $\mathcal{S}_l$:
\begin{theorem}\label{eq:subset_cond}
Suppose we have a solver solving for the MB of a set of random variables and we apply that solver at each layer of a neural network as described in equation (\ref{eq:back_propagate}), then the solution returned by the solver at each layer must contain the MB of the neural network's output at that layer, i.e. $\mathcal{M}_l (Y) \subseteq \mathcal{S}_l, \ \forall l = 0,...,L-1$. \textbf{Proof:}  See Appendix~\ref{proof:lemma}.
\end{theorem}

From Theorem~\ref{eq:subset_cond}, we obtain Corollary~\ref{corollary:solve}, which states any MB solvers with precision guarantee on the input-response pair $(Z_l, Y)$ can be used to solve for the MB of the pair $(Z_l, Z_{l+1}^{\mathcal{S}_{l+1}})$ (eq. (\ref{eq:back_propagate})) with the same precision guarantee:

\begin{corollary}\label{corollary:solve}
 Suppose we have a solver solving for the MB of a random response $T$ with the precision at least $p$ for a given $0<p<1$. Let $\hat{\mathcal{M}}_l$ be the output of that solver on the input-response pair $(Z_l, Y)$ defined in procedure (\ref{eq:back_propagate}). Then, $\hat{\mathcal{M}}_l$ also satisfies the precision guarantee $p$ as if we solve for the input-response pair $(Z_l, Z_{l+1}^{\mathcal{S}_{l+1}})$. \textbf{Proof:}  See Appendix~\ref{proof:corollary}.
\end{corollary}

 Corollary~\ref{corollary:solve} enables us to exploit any solver with precision control to efficiently solve for procedure (\ref{eq:back_propagate}) with precision guarantee.
In our implementation of NeuCEPT-discovery, we use Model-X Knockoffs~\cite{Candes2016}, whose technical aspects are discussed in more details in Appendix~\ref{subsect:modelx}.

 \section{Information-theoretic Interpretation and NeuCEPT-learning of Critical Neurons}\label{sec:learning}
 
 
 The goal of finding critical neurons is to  correctly identify the model's mechanisms. Sect.~\ref{subsec:intuition} discusses in more detail how the MI objective (Eq. (\ref{eq:trueobjective}) in Sect. \ref{subsec:formulation}) is apt for the task. Sect.~\ref{subsect:learning} describes how NeuCEPT extracts information from critical neurons to identify the model's mechanism.
 
 \subsection{Information-theoretic Interpretation} \label{subsec:intuition}

\begin{figure}[ht]
			\centering
			\includegraphics[width=0.85\linewidth]{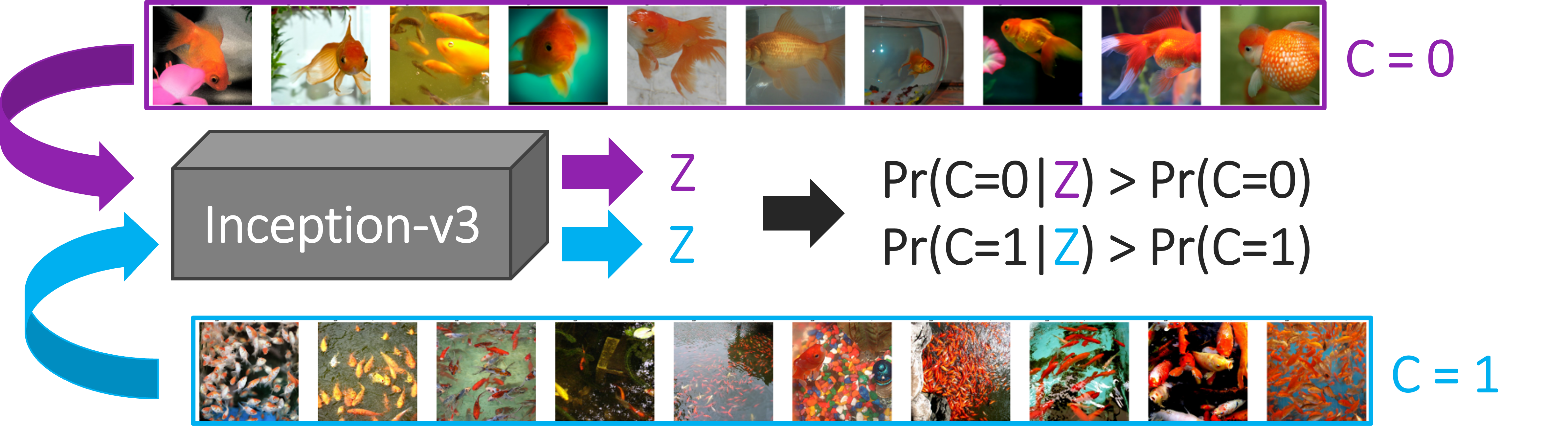}
			\caption{{The probabilistic intuition of the neurons' activation and mechanisms: the mechanism determines the activation, and, conversely, observing the activation can reveal the mechanism.}}
			\label{fig:intuition}
\end{figure}

  \textbf{Mechanism and Explainability power.}  Previous analysis of DNNs~\cite{Bau2020} and our examples (Figs.~\ref{fig:chexnet} and~\ref{fig:fishbee}) reveal distinctive patterns of neurons' activation shared among some input samples. This similarity suggests they might be processed in the same manner by the model, which is what we call \textit{mechanism}. We approach mechanism empirically: similar to how unlabeled data is handled in unsupervised learning, mechanism is modeled as a discrete latent random variable whose realization determines how the predictions are generated. Specifically,
  Fig.~\ref{fig:intuition} provides an intuition on the relationship between the neurons' activation and mechanisms under this assumption. Suppose the latent mechanism variable $C$ determines the generation of predictions of the class \textit{goldfish} in the Inception-v3. Different realizations of $C$, i.e. $0$ or $1$, result in different patterns in the neurons' activation $Z$. On one hand, these patterns specify how the model predicts, which is the intuitive meaning of mechanism. On the other hand, observing the activation on some neurons, i.e. critical neurons, can be sufficient to determine the realization of $C$, i.e. the model's underlying mechanism\footnote{Observing the input images in this experiment suggests that there might exist a close connection between the mechanisms and the visual concepts, which are \textit{a single goldfish} and \textit{a shoal of fish}. However, examining this connection is not the scope of this work.}. This imposes a necessary condition on the selection of critical neurons: their activation should determine (or reduce significantly the uncertainty of) the mechanism. We call this condition \textit{explainability power}.
  
We can see that the objective (\ref{eq:trueobjective}) fits into the notion of explainability power by considering its solution ${\displaystyle \left\{ \mathcal{S}_l \right\}_{l=0}^{L-1}}$ described in (\ref{eq:back_propagate}). From the definition of the MB, for any set of neurons $ \mathcal{R}_l$ at a layer $l$ that is disjoint with  $\mathcal{S}_l$, we have $Z^{\mathcal{R}_l}_l$ is independent with $Z^{\mathcal{S}_{l+1}}_{l+1}$ given $Z^{\mathcal{S}_l}_l$. Since $Z^{\mathcal{S}_{l+1}}_{l+1}$ determines $Z^{\mathcal{S}_{l+2}}_{l+2}$, variables in $\mathcal{R}_l$  must also independent with $Z^{\mathcal{S}_{l+2}}_{l+2}$ given $Z^{\mathcal{S}_l}_l$. For the same reason, we have variables in $\mathcal{R}_l$ independent with all the later variables that contribute to the examined prediction; thus, knowing $Z^{\mathcal{R}_l}_l$ does not provide any more information on how the model generates the prediction of interest. Thus, ${\displaystyle \left\{ \mathcal{S}_l \right\}_{l=0}^{L-1} }$ is sufficient.

\textbf{Non-redundancy.} Neurons' activation not only determine the mechanism but also contain other distinctive patterns. For example, unsupervised learning on MNIST's images, which can be considered as activation of neurons at an input layer, can classify the data into clusters of digits. However, not all patterns must be used by the model to generate predictions. Even if the model learns the patterns, it might exploit them to generate the predictions of other classes, not the class of our examination. Thus, the capability to differentiate inputs based on  neurons' activation does not imply the model's mechanisms. In short, the \textit{non-redundancy} requires the identified mechanisms must be used by the model in generating its prediction. In Appendix~\ref{appendix:nonredundancy}, we explain how our proposed objective (\ref{eq:trueobjective}) fits for the non-redundancy requirement by comparing it to another objective solving for globally important neurons.

\renewcommand\thesubfigure{\normalsize{Figure \arabic{subfigure}:}}
 \begin{figure*}[t]
         \begin{subfigure}[t]{.32\linewidth}
         \addtocounter{subfigure}{4}
         \centering
    	\includegraphics[width=0.96\linewidth]{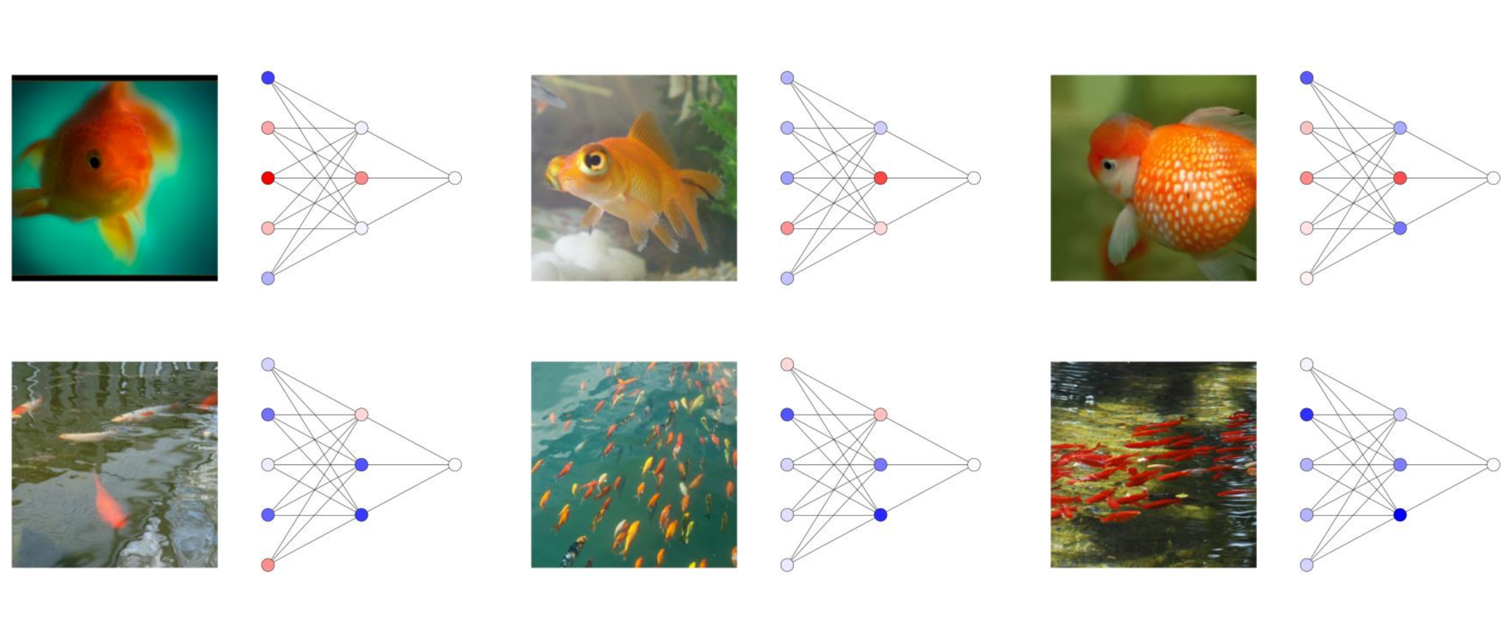}
		\caption{\small{Examples of NeuCEPT's outputs on Inception-v3. Images from each row are from the same cluster learnt by NeuCEPT.}}
		\label{fig:examples}
         \end{subfigure} \ 
    \begin{subfigure}[t]{.67\linewidth}
    \centering	
    \includegraphics[width=0.96\linewidth]{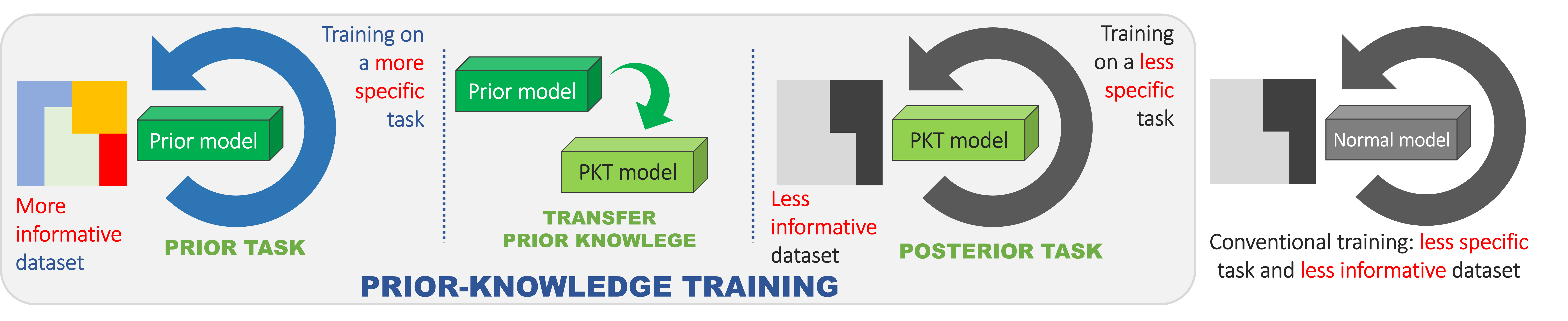}
		\caption{\small{Prior-knowledge training: parameters of the prior-model trained on a more informative task are transferred to the PKT model, which is later trained on a less specific task. The mechanisms of the PKT model is expected to be different from normally-trained models.}}
		\label{fig:priortraining}
		\end{subfigure}
\end{figure*}
\addtocounter{figure}{1}

\begin{figure*}[ht]
    \centering
    \includegraphics[width=0.89\linewidth]{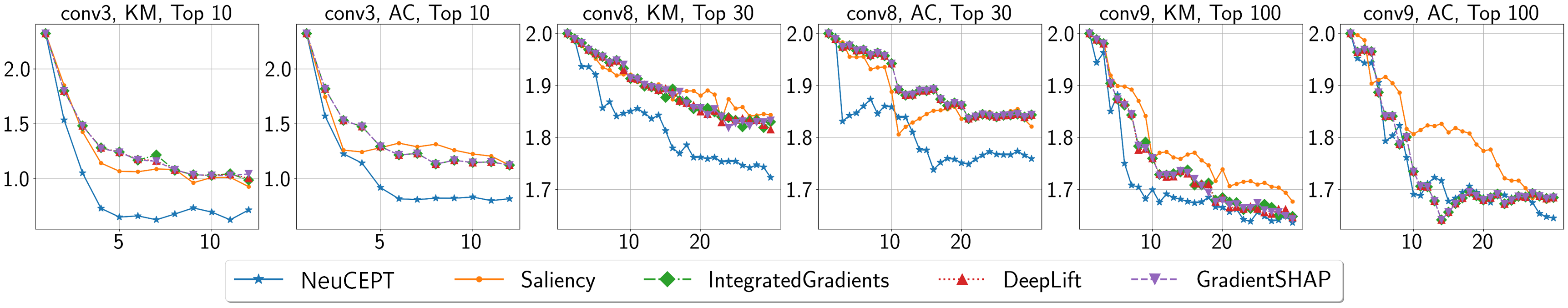}
    \caption{CE (bits) resulted from k-mean (KM) and Agglomerative clustering (AC) clustering on activation of neurons selected by different methods. The x-axis and y-axis are the number of clusters and the CE, respectively. The first 2 plots are the analysis at layer \textit{conv3} of the LeNet on the class \textit{even} using top-10 neurons. The last 4 plots are of VGG on class \textit{object} with top 30/100 neurons at layer \textit{conv8} and \textit{conv9}.}
    \label{fig:distill}
\end{figure*}

\subsection{NeuCEPT-learning on Critical Neurons} \label{subsect:learning}
Given a set of critical neurons identified by NeuCEPT-discovery, 
the goal of NeuCEPT-learning is to extract information from them to identify model's mechanisms. Since the ground-truth mechanisms are not given, it is natural to consider the mechanism identification (or mechanism discovery) problem as an unsupervised learning task, which has been extensively studied~\cite{Goodfellow-et-al-2016}.

Algorithm~\ref{algo:unsupervised_learning} (Appendix~\ref{sect:algo}) describes how NeuCEPT extracting information from critical neurons. Besides their activation, the algorithm's inputs include a set of compactness parameters limiting the number of representative neurons and an integer $K$ defining the target number of mechanisms. The usage of the compactness parameters are commonly used by existing explanation methods for visualization purpose.

The first step of NeuCEPT-learning imposes the compactness requirement. It either selects the top relevant neurons identified by NeuCEPT-discovery or aggregates critical neurons at each layer into a smaller set of representative neurons using feature agglomeration. Similar techniques have been used to apply Model-X Knockoffs on real-data with very high-correlated features~\cite{Candes2016}. Next, we apply one unsupervised learning method among K-means, Gaussian Mixture, and Agglomerative Clustering to map each input sample to one of the $K$ clusters representing $K$ mechanisms. 

We end this section with a demonstration of some NeuCEPT's outputs in analyzing predictions of Inception-v3. The examined layers are the last layers of the \textit{Mixed-5d} and the \textit{Mixed-6e} blocks (Fig. 5). The number of representative neurons are restricted to 5 and 3. Next to each input, we show a graph representing the activation's level (red for high, blue for low) of those representative neurons from the \textit{Mixed-5d} (left) to the \textit{Mixed-6e} (middle). The last dot represents the output neuron (right). NeuCEPT-learning helps us visualize similar activation's patterns among samples of the same mechanism, and differentiate them from another mechanism. 

\begin{figure*}[ht]
    \centering
    \includegraphics[width=0.89\linewidth]{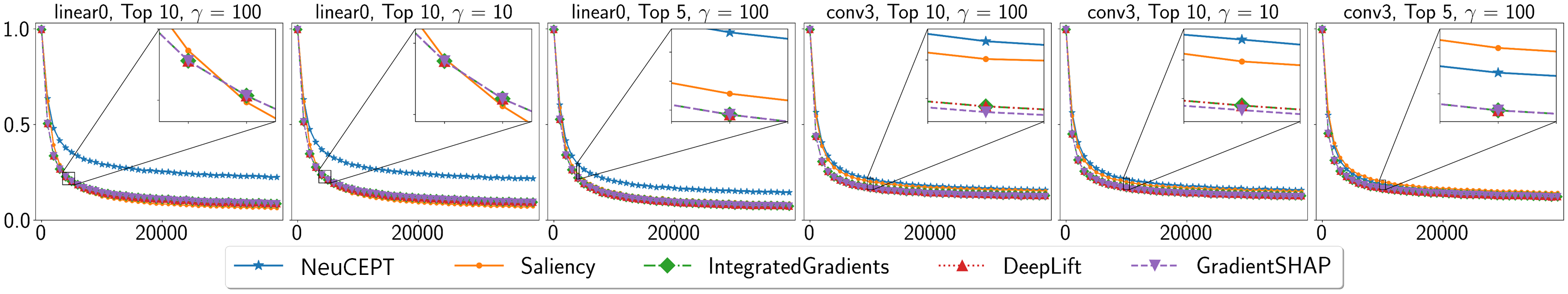}
    \caption{Ablation tests on LeNet. The x-axis and y-axis are the noise levels and the test accuracy. Parameters are specified in Appendix~\ref{appendix:ablation}.}
    \label{fig:ablation_small}
\end{figure*}
\begin{figure*}[ht]
    \centering
    \includegraphics[width=0.89\linewidth]{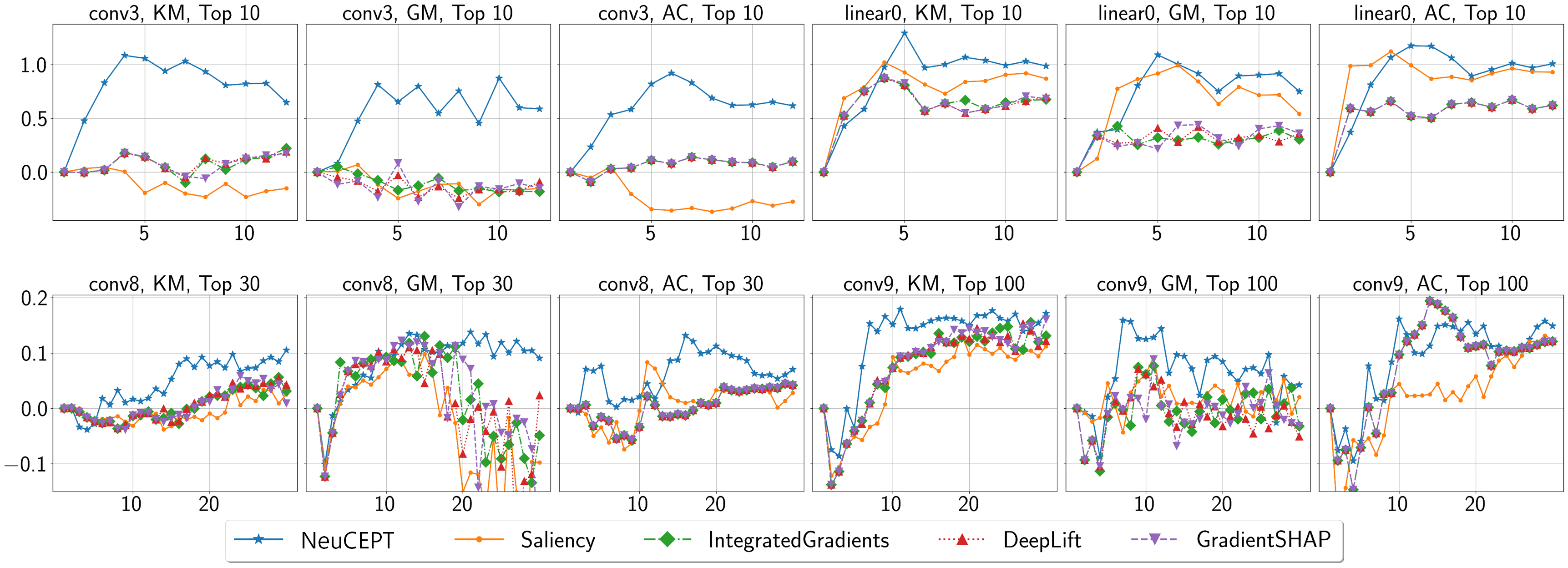}
    \caption{The CE differences (bits) between the conventionally trained model and the PKT model. Clusters are learnt using k-mean (KM), Gaussian Mixture (GM), and Agglomerative clustering (AC) on activation of neurons chosen by different methods. The x-axis and y-axis are the number of clusters and the CE differences. Top figures show the results of the LeNet at different layers with top 10 important neurons. Bottom are the results of VGG with top 30/100 neurons at convolutional layer 8 and 9.}
    \label{fig:cluster}
\end{figure*}

\section{Experiments: Setting and Results} \label{sec:experiments}
Our experiments focus on evaluating the explainability power and the non-redundancy properties, mentioned in Sect.~\ref{subsec:intuition}. While evaluating the explainability power can be conducted via ablation study~\cite{Nguyen2015DeepNN}, evaluating the non-redundancy is more challenging as we normally do not know the underlying mechanism. To tackle this, we propose a training setup, called \textit{prior-knowledge training}, so that both the non-redundancy and the explainability power can be evaluated.

\textbf{Prior-knowledge training (PKT).} The goal of PKT is to inject prior-knowledge into a model, called PKT model, before its training on its main task, called \textit{posterior-task}. Due to the injection, certain mechanisms are expected to be embedded and used by the PKT model to conduct the posterior-task. Fig. 6 describes the process: First, a model, called \textit{prior-model}, is trained on a more specific task, called \textit{prior-task}, with a more informative dataset. The prior-model's parameters are then transferred to the PKT model, which is subsequently trained on the less specific posterior-task (the main task) on a less informative dataset. The prior-task's training data is setup to be more informative so that there is exclusive information only available to the prior-model. The prior-task is set to be more specific because we want the PKT model contains more distinctive mechanisms. In fact, given a PKT model and a conventionally trained model, a good mechanism discovery algorithm should conclude that the PKT model relies on some information, i.e. the prior-knowledge, to generate its prediction (explainability power) while the conventionally trained one does not (non-redundancy). 
As such, we evaluate the algorithms using the clusters' entropy (CE) metric:
\begin{align*}
     \sum\nolimits_{c}  \sum\nolimits_{y_{prior}}  p(y_{prior}, c) \log \left({1}/{p(y_{prior}| c)}\right),
\end{align*}
where $c$ is the identified cluster, $y_{prior}$ is the label from the informative dataset and the $p(.)$ is the empirical probability computed on the data of testing. Intuitively, the lower the CE, the more the clusters/mechanisms identified by the algorithm align with the prior-knowledge. Thus, the CE of PKT model resulted from a good mechanism discovery algorithm should be lower than that of a conventionally trained model.
The details of our experiments, i.e. baseline and implementations, are specified in Appendix~\ref{appendix:experiment}.

\textbf{Explainability power.} The explainability power implies the selected neurons should hold information to determine the mechanism. Thus, the CE should be small if the number of the clusters $K$ (specified in Sect.~\ref{subsect:learning}) aligns with the number of actual mechanisms. In the posterior-tasks, we train LeNet and VGG to classify \textit{even/odd} on MNIST and \textit{animal/object} on CIFAR10, respectively. The numbers of actual mechanisms in the PKT models are expected to be at least $(5,5)$ and $(4,6)$ for each posterior-tasks' label, as they are the number of original labels belonging to those category, i.e. $5$ odd digits.   

Fig.~\ref{fig:distill} shows the CE of the clusters learnt on activation of neurons identified by explanation methods at different $K$. Experiments on MNIST clearly show that neurons identified by NeuCEPT can differentiate the inputs based on their original labels, whose information is embedded in prior-knowledge training. Notably, when $K=5$, NeuCEPT achieves its lowest, which aligns with our expectation on the number of actual mechanisms that the PKT model uses. Similar results can also be observed in VGG, but with less distinction in the number of clusters. Since the dataset and the models are both much more complicated, an explanation of this behavior is that the number of actual mechanisms might be much larger than that in previous experiments. Additionally, Fig.~\ref{fig:ablation_small} shows our ablation tests of neurons identified by different methods on LeNet. The results show that neurons identified by NeuCEPT hold higher predictive power among most of the experiments, which implies their strong explainability power. Further ablation results and details are reported in Appendix~\ref{appendix:ablation}.

\textbf{Non redundancy.} 
This set of experiments shows that NeuCEPT does not return the same CE when analyzing a conventionally trained model, which has no knowledge on the original labels. Fig.~\ref{fig:cluster} plots the differences in the CE between a conventionally-trained model and the PKT model on both MNIST and CIFAR10. While the differences in CE at certain layers of some other methods fluctuate around $0$, indicating there is no difference between the two models, NeuCEPT consistently supports the expectation that only the PKT model recognizes the prior-knowledge. Note that NeuCEPT can differentiate the two models trained differently simply by observing their activation while other methods fail to do so (see the 3 top-left plots of  Fig.~\ref{fig:cluster} as examples).


\section{Conclusion} \label{sec:conclude}
In this paper, we attempt to discover different mechanisms underlying DNNs' predictions in order to provide a deeper explanation on how a model works. Based on an information-theoretic viewpoint, we formulate the problem as a sequence of MI maximization. The solution of the maximization, called critical neurons, can be solved by our proposed algorithm NeuCEPT-discovery with precision guarantee. Then, we propose NeuCEPT-learning, an algorithm clustering inputs based on their activation on critical neurons so that the underlying model's mechanism can be revealed. We also propose a prior-knowledge training procedure so that the mechanism discovery task can be evaluated. Our experimental results show that NeuCEPT is able to consistently identify the underlying mechanism of DNNs' predictions. 

\clearpage
\bibliographystyle{named}
\small{
\bibliography{ijcai22}
}

\renewcommand\thesubfigure{(\alph{subfigure})}

\clearpage
\appendix

\section{Illustrative examples for motivations} \label{appendix:motivation}
This appendix contain illustrative examples showing the benefits of identifying critical neurons and obtaining knowledge on the model's mechanisms. First, we show how our research can contribute to the study of model's linear separability (Sect.~\ref{appendix:linearprobe}). Then, we provide illustrative example showing how knowledge on the model's mechanisms help us re-evaluate the model's outputs 
(Sect.~\ref{append:chexnet}). We also demonstrate that simply selecting class-highly activated neurons, which can be considered as a simple feature-attribution method, would miss some neurons holding valuable information on the model's mechanisms (Sect.~\ref{append:inception}). 

\subsection{Example on Linear Probe}\label{appendix:linearprobe}

Linear Probes~\cite{alain2017understanding} are linear classifiers hooking on intermediate layers of neural networks to measure their linear separability. Intuitively, lower the loss of the classifier at a layer suggests that the activation of neurons in that layer is more linearly separable. This metric is important and interesting in the analysis of neural networks since it can be used to characterize layers, to debug models, or to monitor training's progress.

\begin{figure}[ht]
     	\centering
			\includegraphics[width=0.8\linewidth]{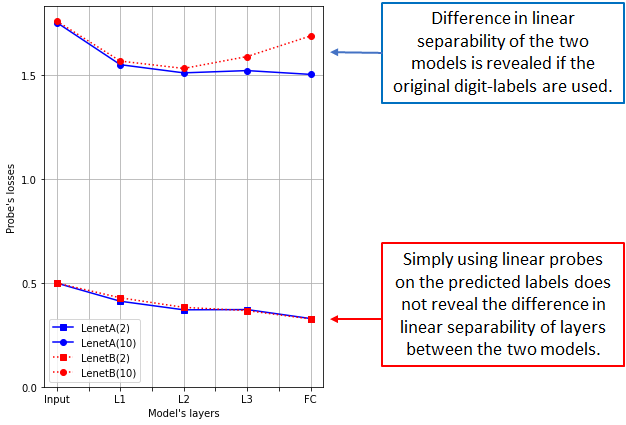}
		\caption{Simply using linear probes on the predicted labels, i.e. \textit{even/odd}, does not reveal the difference in the linear separability of LeNetA and LeNetB (indicated by notation (2)). However, using the knowledge on the original labels of the dataset, linear probes show that the linear separability of the two models are difference (indicated by notation (10)).}
		\label{fig:linearprobe}
 \end{figure}
 
We applied Linear Probes to study LeNet(A) and LeNet(B) (Fig.~\ref{fig:dnn}) on their \textit{even/odd} labels and plotted the results in Fig.~\ref{fig:linearprobe} with notation LeNetA(2) and LeNetB(2), respectively. The result shows that there is little difference between the linear separability between layers of the two models. However, as shown in the visualization of the point-clouds identified by critical neurons (bottom Fig.~\ref{fig:dnn}), we expect that the degree of linear separability of the two model should be different. Specifically, LeNet(A) is expected to be more linearly separable since its point-cloud is more separated into smaller clusters. Indeed, Fig.~\ref{fig:linearprobe} also demonstrates if we use Linear Probes to measure how well the layers' activation can predict the original digit labels of the dataset, there is a clear difference in the corresponding results (indicated by notation LeNetA(10) and LeNetB(10)). 

The results indicate two major meanings. First, it shows that information obtained by NeuCEPT is consistent with Linear Probe, i.e. they both suggest LeNet(A) is more linearly separable, and they can be used to complement each other. Secondly, in practice, we normally do not have the labels on the model's mechanisms; thus, the gap like that between LeNetA(10) and LeNetB(10) cannot be obtained by Linear Probe. This means the linear separability might not be captured correctly by Linear Probe (as shown in the case of LeNetA(2) and LeNetB(2)). In that situation, NeuCEPT can be used to ensure the results obtained by Linear Probe or further the study of linear separability.

\begin{figure*}
    \centering
    \begin{subfigure}[c]{.72\linewidth}
        \centering
        \includegraphics[width=0.99\linewidth]{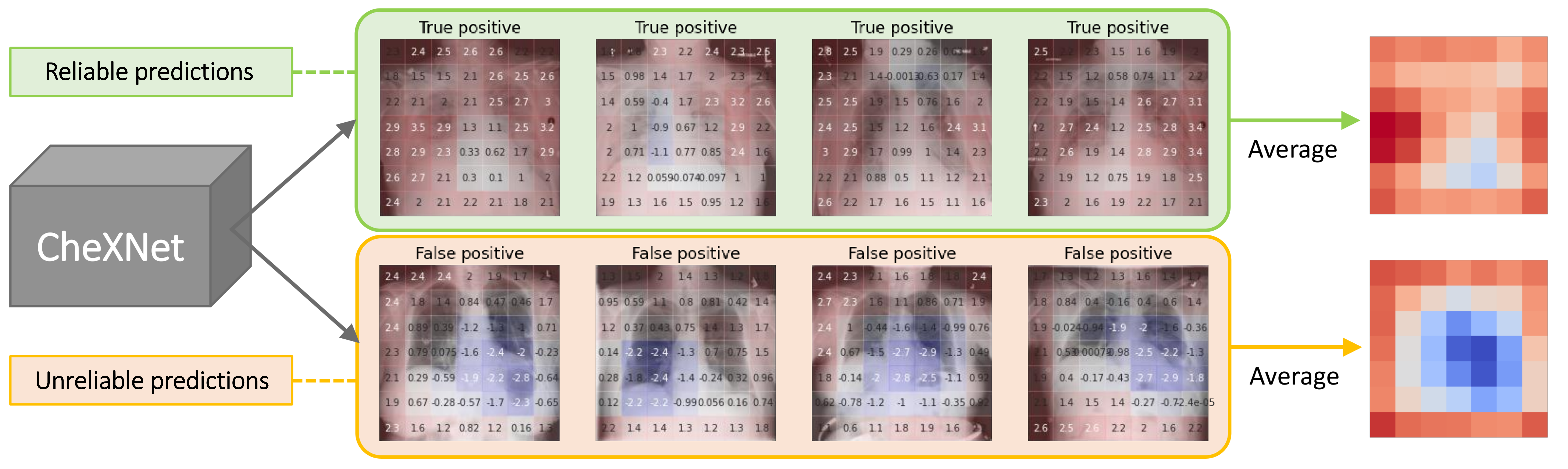}
    \end{subfigure}
    \begin{subfigure}[c]{.27\linewidth}
        \centering
                    \scalebox{0.52}{
                    \begin{tabular}{@{}ll@{}}
                    \multicolumn{2}{l}{\textbf{Experiment's info}}                                                                                                           \\ \midrule
                    \multicolumn{1}{l|}{Model}                                                                & CheXNet                                                           \\ \midrule
                    \multicolumn{1}{l|}{Examined predictions}                                                 & \textit{'Pneumonia'}                                              \\ \midrule
                    \multicolumn{1}{l|}{Total samples}                                                        & 2978                                                              \\ \midrule
                    \multicolumn{1}{l|}{Positive samples}                                                     & 1035                                                              \\ \midrule
                    \multicolumn{1}{l|}{Positive predictions}                                                 & 1035                                                              \\ \midrule
                    \multicolumn{1}{l|}{Model's precision}                                                    & $60.6\%$                                                          \\ \midrule
                    \multicolumn{1}{l|}{\begin{tabular}[c]{@{}l@{}}Reliable-subclass \\ precision\end{tabular}}   & \begin{tabular}[c]{@{}l@{}}$91.3\%$ \\ (180 samples)\end{tabular} \\ \midrule
                    \multicolumn{1}{l|}{\begin{tabular}[c]{@{}l@{}}Unreliable-subclass \\ precision\end{tabular}} & \begin{tabular}[c]{@{}l@{}}$36.7\%$ \\ (150 samples)\end{tabular} \\ \bottomrule
                    \end{tabular}
                    }

    \end{subfigure}
    \caption{\small{NeuCEPT discovers a subclass of 
    unreliable predictions with precision of $36.7\%$ (compared to $60.6\%$ on the whole class) in CheXNet. We also extract the explanation heat-maps provided by the model which suggest that the unreliable positive predictions are made by the area at the boundary of the image instead of the lung's area (red means high importance score).}}  
    \label{fig:chexnet}
\end{figure*}
\begin{figure*}
		\centering
		\begin{subfigure}[t]{0.48\linewidth}
		\centering
			\includegraphics[width=0.95\linewidth]{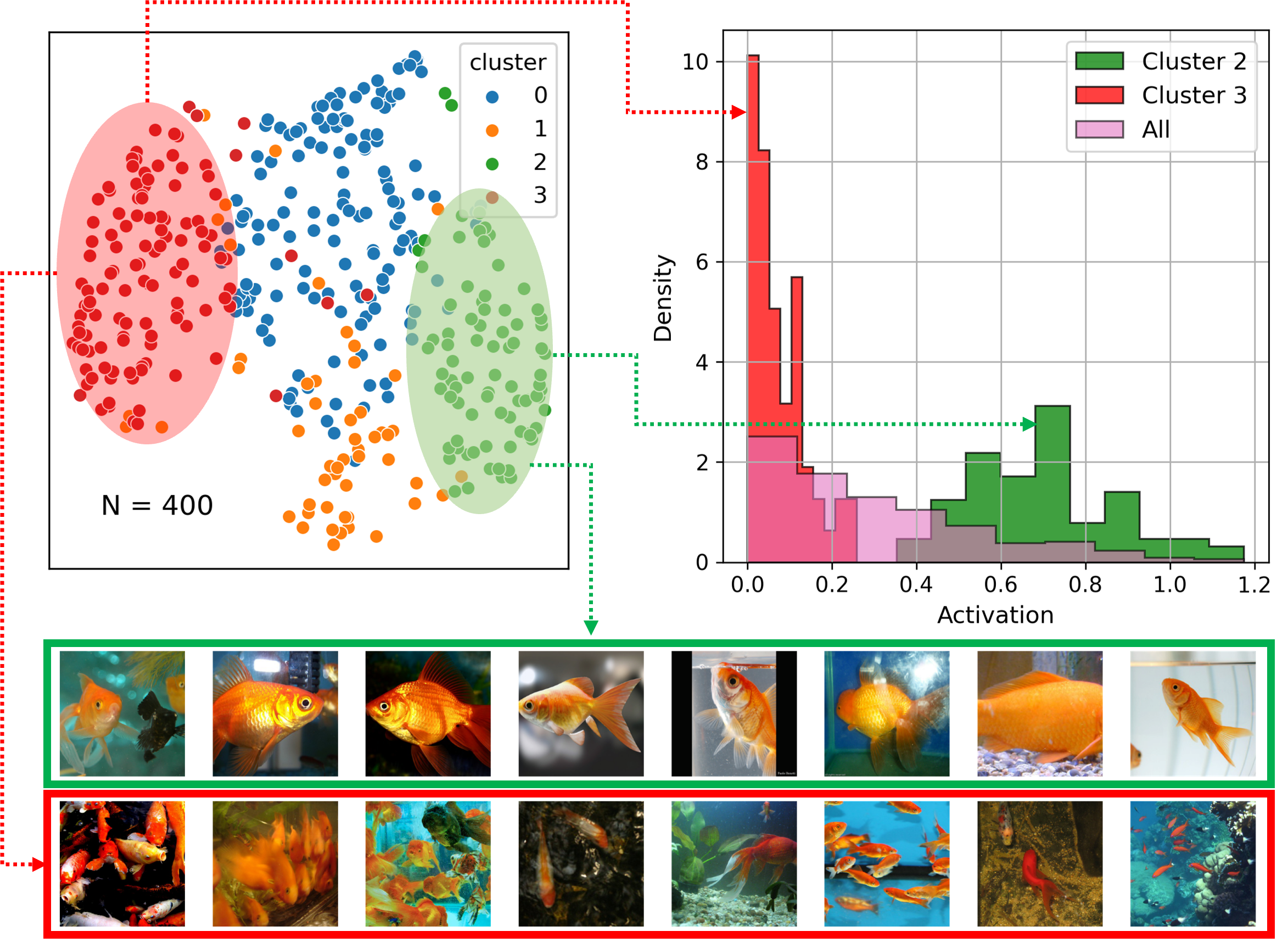}
        \caption{Analysis on class \textit{goldfish} at the \textit{Mixed6e} block of Inception-v3 indicates neuron 322 holds valuable information in differentiating predictions of two sub-classes (red and green).}
		\label{fig:fishRational}
		\end{subfigure} \
	    \begin{subfigure}[t]{0.48\linewidth}
	    \centering
			\includegraphics[width=0.95\linewidth]{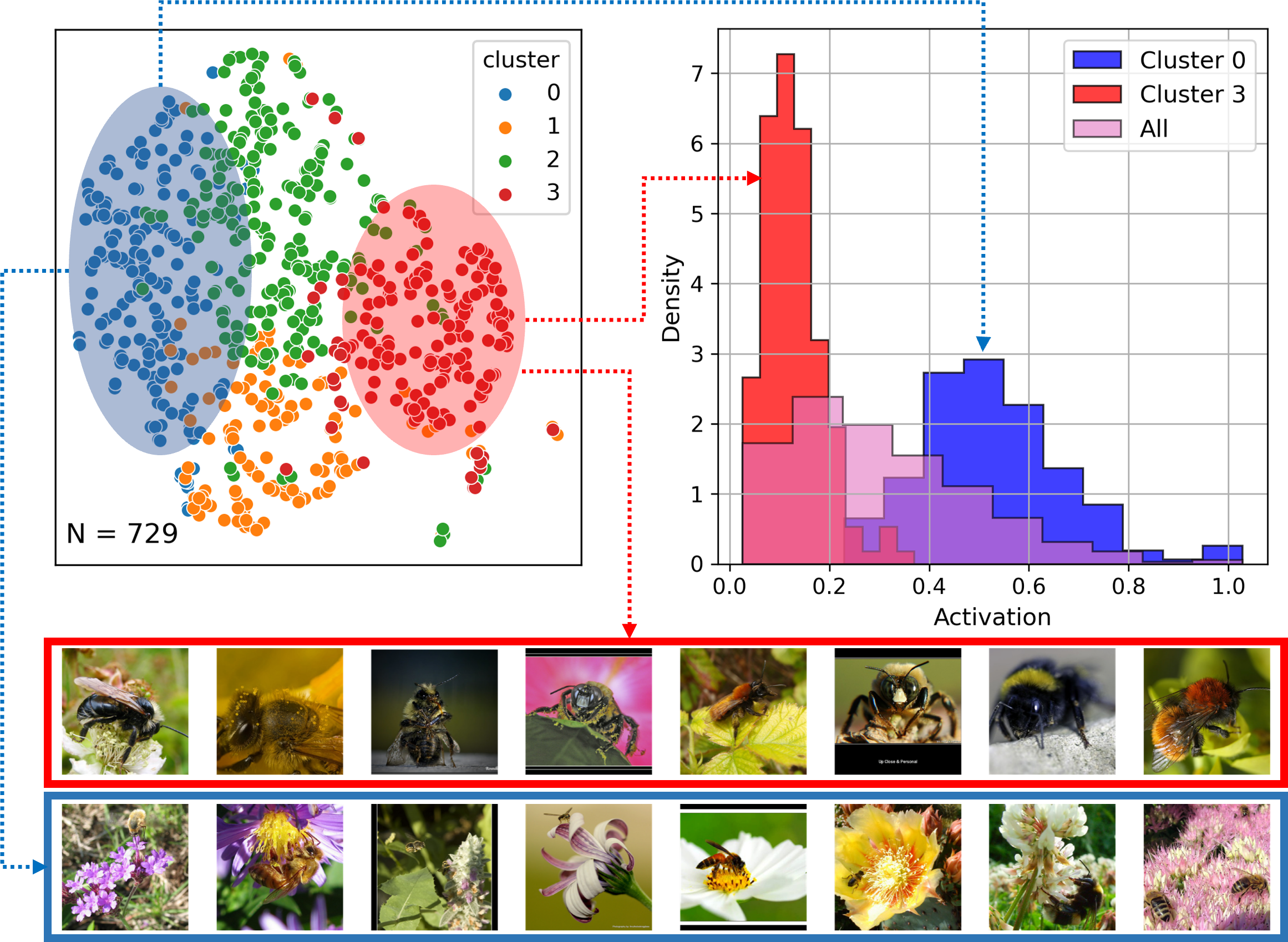}
        \caption{Analysis on class \textit{bee} at the \textit{Mixed6e} block of Inception-v3 indicates neuron 305 holds valuable information in differentiating predictions of two sub-classes (blue and red).}
		\label{fig:beeRational}
		\end{subfigure}
		\caption{Results obtained by applying NeuCEPT with the number of clusters $K =4$ on classes \textit{goldfish} and \textit{bee} at the last layer of the \textit{Mixed6e} block of Inception-v3. For each class, NeuCEPT identifies a neuron whose activation's histogram on two distinctive clusters (among four) learnt by NeuCEPT and all inputs are shown on the top-right figure. While these neurons are not among the top-activated neurons of each class, they hold vital information in differentiating the predictions as inputs of one cluster activate the neuron and the others do not. The top-left and the bottom figures plot the low-dimensional representations of the inputs and some images drawn from each cluster, respectively, for visualization.}
		\label{fig:fishbee}
\end{figure*}

\subsection{Example on CheXNet}\label{append:chexnet}

In examining CheXNet~\cite{CheXNet} using NeuCEPT, we discover a subclass of unreliable \textit{'pneumonia'} predictions with a much lower precision than that on the whole class, i.e. $36.7\%$ compared to $60.6\%$.  
The results are described in Fig.~\ref{fig:chexnet}. Similar unreliability's behavior is also observed in previous works\footnote{In the work "Variable generalization performance of a deep learning model to detect pneumonia in chest radio-graphs: A cross-sectional study", \textit{PLoS Medicine} (2018), Zech et al. obtained activation heat-maps from a trained CheXNet and averaged them over a set of inputs to reveal which sub-regions tended to contribute to a classification decision. They found out that the model has learned to detect a hospital-specific metal token on the scan to generate some classifications.}. Furthermore, same conclusion can be deducted based on the local explanations provided by the model itself, i.e. the false-positive predictions generally are made using features outside of the lung's area as shown in the figure. 

This example clearly shows that end-users need systematical methods to differentiate reliable predictions from unreliable ones. Identifying mechanisms underlying the model's predictions is therefore vital since it can help divide predictions into groups of similar mechanisms. Based on the mechanisms, researchers and practitioners can systematically evaluate the reliability of samples.
Note that, without this information, we might need to discard the model and cannot exploit its high predictive power due to some unreliable predictions.

\subsection{Example on Inception-v3}\label{append:inception}

As the main goal of attribution methods is to identify neurons contributing the most to the prediction, aggregating the resulted attribution scores among inputs of a class naturally gives us the neurons highly attributing to the class. However, as the activation of those neurons can be highly similar for almost all samples of the class (one trivial example is the output's neuron associated with the class), they hold limited information on how the model processes those inputs differently, hence, do not fit for the task of identifying mechanisms. 

To illustrate this point, Fig.~\ref{fig:fishbee} provides our analysis on the class \textit{goldfish} and \textit{bee} of Inception-v3~\cite{Inception2015}. At the last layer of the \textit{Mixed6e} block of the model, for each class, we identify a neuron (using NeuCEPT) that is not among the top highly activated of the class, but hold valuable information in determining how the model views inputs of the class differently. Specifically, as shown in the figure, there are two subsets of inputs of the examined class such that inputs in one subset activate the neuron while the others do not. Simply using activation's level, which can be considered as the simplest form of attribution method, would miss this neuron. 

An interesting observation is that, in both classes, there are distinctive visual differences in the images drawn from the clusters identified by NeuCEPT. In examples of \textit{goldfish}, which has been used throughout the main manuscript, samples in one cluster show a single fish while those in the other show a shoal of fishes. On the other hand, in the class \textit{bee}, samples of a cluster show a single bee while the samples in the other mainly focus on a huge flower or a bunch of flowers. This observation supports the hypothesis that the two subsets of images are indeed processed differently by the model.

\section{CNI is in NP-hard}\label{proof:nphard}
The CNI can be considered as a more general version of the feature-selection problem with mutual-information objective, which is known to be NP-hard~\cite{KHAIRE2019}. However, we find that there is no recent works concretely prove the NP-hard result of this feature-selection problem. Hence, for the completeness of our result, we provide the proof that the CNI is in NP-hard in the followings.

We first consider the discrete version of the CNI optimization problem in (\ref{eq:trueobjective}), called DISCRETE-CNI, as follows: Given a random variable $Z$ over the set of binary sequences $ \{0,1\}^N$, a discrete random variable $Y$ and a constant integer $k, 1 \leq k \leq N$, we want to find the set of indices $\mathcal{S}^*$ such that:
\begin{align}
        \mathcal{S}^* = \textup{argmax}_{\mathcal{S} \subseteq \{1,2,...,N\}} I\left(Z^\mathcal{S};Y \right), \textup{ s.t. } |\mathcal{S}| \leq k \label{eq:cni},
\end{align}
where $Z^\mathcal{S}$ indicates the set of components $i$ of $Z$ such that $i \in \mathcal{S}$. It is clear that this discrete version can be reduced to the original CNI problem in (\ref{eq:trueobjective}) in polynomial time. 

We show that {DISCRETE-CNI} is in NP-hard by reducing  {MIN-FEATURES}~\cite{Davies94np} to {DISCRETE-CNI}. The {MIN-FEATURES} is defined as follows: Given a set $X \subseteq \{0,1\}^{N'} \times T$ of samples, whose the first $N'$ components called \textit{features} and the last component called \textit{target}, and an integer $n$, determine whether or not there exists some feature set $\mathcal{V}$ such that:
\begin{itemize}
    \item $\mathcal{V} \subseteq \{1,2,...,N'\}$
    \item $|\mathcal{V}| \leq n$
    \item There exists no two examples in $X$ that have identical values for all the features in $\mathcal{V}$ but have different values for the target.  
\end{itemize}
Note that we only consider the case where the target $T$ can always be determined by the $N'$ features. If that is not the case, the problem is trivial since there is no such $\mathcal{V}$. 

To reduce {MIN-FEATURES} to {DISCRETE-CNI}, we define the random variable $Z$ over $ \{0,1\}^{N'}$ of DISCRETE-CNI as follows:
\begin{align*}
    \textup{Pr} \{ Z = x \} = \left\{\begin{matrix}
    1/ |X|, \ & \forall x \in X[1:N'] 
    \\ 
    0,  & \textup{ otherwise }.
    \end{matrix}\right.
\end{align*}
where $X[1:N']$ is the set of samples of features of $X$, i.e. the set $X$ but without the target. 

We now can reduce MIN-FEATURES by running DISCRETE-CNI on $Z$ as constructed above, $Y$ as the target in $X$ and the cardinality constraint constant $k$ as $n$. We then decide the existence of $ \mathcal{V}$ by checking whether or not the solution $ \mathcal{S}^*$ returned by {DISCRETE-CNI} satisfying $I\left(Z^\mathcal{S^*};Y \right)$ equals to the entropy of $Y$. Specifically, if $ I\left(Z^\mathcal{S^*};Y \right) = H(Y)$, then $ Z^\mathcal{S*}$ determines $Y$ and $\mathcal{S^*}$ satisfies the conditions of {MIN-FEATURES}. On the other hand, if $ I\left(Z^\mathcal{S^*};Y \right) < H(Y)$, then observing features in $\mathcal{S^*}$ does not determine $Y$. This implies there must exist at least two feature-samples of $X[1:N'] $ such that their features in $ \mathcal{S}^*$ are the same but their targets are different. Thus, there is no $\mathcal{V}$ satisfying {MIN-FEATURES} because the existence of such  $\mathcal{V}$ means $\mathcal{S}^*$ is not the optimal of {DISCRETE-CNI}.

\section{Descriptions of NeuCEPT's algorithms} \label{sect:algo}
In this appendix, we provide the descriptions of NeuCEPT-discovery (Alg.~\ref{algo:fdr_ko}) and NeuCEPT-learning (Alg.~\ref{algo:unsupervised_learning}).

\begin{algorithm*}
        \SetAlgoLined 
        \SetKwInOut{Input}{Input}
        \SetKwInOut{Output}{Output}
        \Input{Samples of model's activation $Z = (Z_1,...,Z_M)$  at $M$ given layers. \\
        A set of precision thresholds $p = (p_1,...,p_M)$ for those chosen layers.}
        \Output{ Estimation of critical neurons at all examined layers $ \left \{ \hat{\mathcal{M}_l} \right \}_{l=1}^{M}$.}
        $Y \leftarrow Z^{\{o\}}_{M+1}$ \\
        For {$l = 1$ to $M$} do: \\
        \Indp  $ \hat{\mathcal{M}_l} \leftarrow $ estimation of the Markov blanket of $Y$ at layer $l$ with precision control $p_l$. \\
        \Indm 
        End \\
        Return $ \left \{ \hat{\mathcal{M}_l} \right \}_{l=1}^{M}$.
         \caption{NeuCEPT-discovery:  \\ Critical Neurons Identification.} \label{algo:fdr_ko}
\end{algorithm*}

\begin{algorithm*}
        \SetAlgoLined
        \SetKwInOut{Input}{Input}
        \SetKwInOut{Output}{Output}
        \Input{Critical neurons' activation at $M$ examined layers, denoted as $Z^\mathcal{S}$.  \\
        A set of values limiting the number representative neurons $v = (v_1, \cdots, v_{M})$.\\
        A positive integer $K$ guessing the number of clusters/mechanisms. 
        }
        \Output{The explained-representations of all inputs and their corresponding mechanisms/clusters.}
        $\#$ \textit{Constraints enforcement} \\
        \Indp If $v \neq$ None: \\
            \Indp For {$l = 1$ to $M$} do: \\
            \Indp  $ V_l \leftarrow $ Feature Agglomeration ($Z_l^{\mathcal{S}_l}$) with constraint $|V_l| \leq v_l$ \\
            \Indm 
            End \\
            \Indm 
        Else: \\
            \Indp  $ V \leftarrow $ $Z^{\mathcal{S}}$. \\
            \Indm 
        \Indm
        $\#$ \textit{Unsupervised learning} \\
        \Indp $g \leftarrow $ Initialize an unsupervised model with $K$ components\\
        Fit $g$ on $V = (V_0,\cdots,V_{L-1})$\\
        Explained-representation $E \leftarrow$ transformation of $V$ by $g$\\
        Mechanism $C \leftarrow$ prediction of $V$ by $g$\\
        \Indm 
        Return $E$ and $C$.
        
         \caption{NeuCEPT-learning: \\ Mechanism discovery.} \label{algo:unsupervised_learning}
\end{algorithm*}

\section{Proof of Theorem 1}\label{proof:lemma}
For simplicity, we consider the following Markov chain $Z_0 \rightarrow Z_1 \rightarrow Y$. We now show that
$\mathcal{M}_0(Y) \subseteq \mathcal{M}_0\left(Z_1^{\mathcal{M}_1(Y)} \right)$. We have: 
\begin{itemize}
    \item $Z_1$ determines $Y$ and $Y \independent   \left \{ Z_1 \setminus Z_1^{\mathcal{M}_1(Y)} \right \} | Z_1^{\mathcal{M}_1(Y)} $ so that $ Z_1^{\mathcal{M}_1(Y)}$  determines $Y$. We can also write this conclusion as $Z_1^{\mathcal{S}_1} $ determines $Y$.
    \item $Z_0$ determines $Z_1^{\mathcal{M}_1(Y)}$ and $Z_1^{\mathcal{M}_1(Y)} \independent  \left \{ Z_0 \setminus  Z_0^{\mathcal{M}_0 \left(Z_1^{\mathcal{M}_1(Y)} \right) } \right \} | Z_0^{\mathcal{M}_0 \left(Z_1^{\mathcal{M}_1(Y)} \right) } $ so that $ Z_0^{\mathcal{M}_0 \left(Z_1^{\mathcal{M}_1(Y)} \right) }$ determines $  Z_1^{\mathcal{M}_1(Y)}$. Similar to the previous statement, we can write this conclusion as $Z_0^{\mathcal{S}_0} $ determines $Z_1^{\mathcal{S}_1} $.
    \item Combine the two above statements, we have $ Z_0^{\mathcal{M}_0 \left(Z_1^{\mathcal{M}_1(Y)} \right) }$ determines $Y$.
\end{itemize}

On the other hand, we have $\mathcal{M}_1(Y)$ is the smallest subset of neurons at the $Z_0$ layer that determines $Y$. Due to the uniqueness of the minimal set that separates $Y$ from the rest of the variables (which is the Markov blanket of $Y$)~\cite{PearlPaz_2014}, we have $\mathcal{M}_1(Y) \subseteq \mathcal{M}_1\left(Z_2^{\mathcal{M}_2(Y)} \right)$.

We can see that the proof generalizes for the case of $L$ layers Markov chain $Z_0 \rightarrow Z_1 \rightarrow \cdots \rightarrow Z_L$ as we can apply the same arguments to conclude that $Z_l^{\mathcal{S}_l} $ determines $Z_{l+1}^{\mathcal{S}_{l+1}}$. This would lead to the fact that all $Z_l^{\mathcal{S}_l}$ can determine $Y$; hence, $\mathcal{S}_l$ contains $\mathcal{M}_l (Y)$ due to the uniqueness of the Markov blanket~\cite{PearlPaz_2014}.

\section{Proof of Corollary 1}\label{proof:corollary}
Denote $q = 1 - p$. Since the precision is one minus the false-discovery-rate (FDR), to show the Corollary, we need to prove:
\begin{align*}
     \textup{FDR} \vcentcolon = \mathbb{E} \left[ \frac{\# \{ j: j \in \hat{\mathcal{M}}_l \setminus \mathcal{S}_l \}}{\# \{ j: j \in \hat{\mathcal{M}}_l\}} \right] \leq q
\end{align*} 

 From Theorem~\ref{eq:subset_cond}, we have $\mathcal{M}_l (Y) \subseteq \mathcal{S}_l$ for all $l = 0,\cdots,L-1$. This implies:
 \begin{align}
      &\hat{\mathcal{M}}_l \setminus \mathcal{S}_l \subseteq  \hat{\mathcal{M}}_l \setminus \mathcal{M}_l(Y) \nonumber \\ \Longrightarrow&  \# \{ j: j \in \hat{\mathcal{M}}_l \setminus \mathcal{S}_l \} \leq  \# \{ j: j \in \hat{\mathcal{M}}_l \setminus \mathcal{M}_l(Y) \} \label{eq:setsmaller}
 \end{align}
 On the other hand, as $\hat{\mathcal{M}}_l$ is the solution of the solver on the input-response pair $(Z_l, Y)$ with FDR less than or equal to $q$, we have
 \begin{align}
     \mathbb{E} \left[ \frac{\# \{ j: j \in \hat{\mathcal{M}}_l \setminus \mathcal{M}_l(Y) \} }{ \# \{ j: j \in \hat{\mathcal{M}}_l\}}\right] \leq q. \label{eq:fdr_my}
\end{align}
Combining (\ref{eq:setsmaller}) and (\ref{eq:fdr_my}), we have the Corollary:
\begin{align}
    &\mathbb{E} \left[ \frac{\# \{ j: j \in \hat{\mathcal{M}}_l \setminus \mathcal{S}_l \} }{ \# \{ j: j \in \hat{\mathcal{M}}_l\}}\right]  \nonumber \\
    \leq&
    \mathbb{E} \left[ \frac{\# \{ j: j \in \hat{\mathcal{M}}_l \setminus \mathcal{M}_l(Y) \} }{ \# \{ j: j \in \hat{\mathcal{M}}_l\}}\right] \leq q.
\end{align}

\section{Model-X Knockoffs} \label{subsect:modelx}
\textbf{Background.}
Model-X Knockoffs~\cite{Candes2016} is a new statistical tool investigating the relationship between a response of interest and a large set of explanatory variables. From a set of hundreds or thousands variables, Model-X Knockoffs can identify a smaller subset potentially explaining the response  while rigorously controlling the false discovery rate (FDR). Specifically, Model-X Knockoffs  considers a very general conditional model, where the response $T$ can  depend in an arbitrary fashion on the covariates $R = (R_1,\cdots , R_p)$. The method assumes no knowledge of the conditional distribution of $T|R$ but it assumes the joint distribution of the covariates $R$ is known. The goal of Model-X Knockoffs is to find important components of $R$ that affect $T$, i.e. the members of the Markov blanket for $T$, while controlling a Type I error. In other words, the method aims to select as many variables as possible while limiting the false positives. Denoting $\mathcal{R}$ and $\hat{\mathcal{R}}$ the Markov blanket and the outcome of the Model-X Knockoffs procedure, the  FDR is:
\begin{align}
    \textup{FDR} = \mathbb{E} \left[ \frac{\# \{ j: j \in \hat{\mathcal{R}} \setminus \mathcal{R} \} }{ \# \{ j: j \in \hat{\mathcal{R}}\}}\right],
\end{align}
with the convention $0/0 = 0$. For each sample of ${R}$, Model-X Knockoffs generates a knockoff copy $\Tilde{R}$ satisfying $Y \independent \Tilde{R} | R$ and the \textit{pairwise exchangeable} property~\cite{Candes2016}. Then, some importance measure $U_j$ and $\Tilde{U}_j$ are computed for each $R_j$ and $\Tilde{R}_j$, respectively. After that, the statistics $W_j = U_j - \Tilde{U}_j$ is evaluated for each feature where a large and positive value of $W_j$ implies evidence against the hypothesis that the $j^{\textup{th}}$ feature is not in the Markov blanket of $T$. The works \cite{Candes2016} have shown that exact control of the FDR below the nominal
level $q$ can be obtained by selecting $\hat{\mathcal{R}} = \{ j: W_j \geq \tau_q\}$, with
\begin{align}
    \tau_q = \min \left \{ 
    t > 0 : \frac{1 + | \{ j: W_j \leq t \} | }{| \{ j: W_j \geq t \}  |} \leq q
    \right \},
\end{align}
where the numerator in the above expression can be considered as a conservative estimate of the number of false
positives above the fixed level $t$. In this manuscript, we use the tuple $(.,.)$ to denote the input-response pair for Model-X Knockoffs as a general solver for the Markov blanket. For example, the input-response pair of the formulation above is $(R,T)$.

\textbf{Technical implementation.}
There are several concerns and details the usage of Model-X Knockoffs on high-dimensional real-data and, specifically, on neurons' activations. One of them is the problem that it is very difficult to choose between two or more very highly-correlated features. Note that "this is purely a problem of power and would not affect the Type I error control of knockoffs"~\cite{Candes2016}. To overcome this challenge, \citeauthor{Candes2016} clustered
features using estimated correlations as a similarity measure, and discovered important clusters. After that, one representative is chosen from each cluster. The representatives are then used in the feature-discovery task instead of the original features. Our feature agglomeration step (described in Sect.~\ref{sec:learning} and Algo.~\ref{algo:unsupervised_learning}) is based on this alleviation. 

Another concern on the usage of Model-X Knockoffs is an unavailability of the true joint covariate distribution, which forces the user to estimate it from the data. The usage of NeuCEPT falls into this case since the neuron's activation variable $Z$ can only be explored using available input samples. This raises the question on how robust NeuCEPT is to this error introduced by Model-X Knockoffs. Theoretical guarantees of robustness are not the scope of this paper; however, \citeauthor{Candes2016} has extensively provided testings on both synthetic and real-world data to evaluate the robustness of Model-X Knockoffs in their work. The main observations are the FDR is rarely ever violated even for a loosely estimated covariance, and the power of selected features only lost about $20\%$ as if they would have had with an exact covariance.

\section{Discussion on the non-redundancy} \label{appendix:nonredundancy}
To demonstrate how the optimal solution of (\ref{eq:trueobjective}), $\mathcal{S}_l$, which is approximated by NeuCEPT-discovery, meets the notion of non-redundancy, we compare it with solution of another objective aiming to identify the set of {globally important} neurons in a given layer $l$. Intuitively, they are more about all the predictions of the model than some specific predictions. From the Markov chain (\ref{eq:markov}), a natural selection of those globally important neurons can be formalized as:
\begin{align}
        \mathcal{S}^*_l = \textup{argmax}_{\mathcal{S} \subseteq \mathcal{N}_l} I\left(Z_l^{\mathcal{S}};Z_{l+1}, ..., Z_L \right), \textup{ s.t. } \mathcal{S} \in \mathcal{C} \label{eq:objective}
\end{align}
where the notations have the same meaning as those in equation (\ref{eq:trueobjective}). This objective tells us how much we know about the model's later layers given the activation of the neurons in $\mathcal{S} \subseteq \mathcal{N}_l$. We denote $\Phi_0 (Z)$ and $\Phi_1 (Z)$ as the sums of the objectives (\ref{eq:objective}) and  (\ref{eq:trueobjective}), respectively. We then naturally have $\Phi_0 (Z) \geq \Phi_1 (Z)$. From the lens of information theory, $\Phi_0(Z)$ is the sum of the minimum description length of the model's layers, i.e. information encoded in those neurons can fully specify the model as well as its layers. However, when examining the prediction on a specific class, the information in ${\displaystyle \left \{\mathcal{S}^*_l \right \}_{l=0}^{L-1}}$ is too excessive since the objective does not consider the classes differently. Therefore, mechanisms discovered on top of them is prone to redundancy. This reason motivates us to consider $\mathcal{S}_l$ and the objective (\ref{eq:trueobjective}) for our study. This also implies that the critical neuron discovery step in NeuCEPT is necessary.

\begin{table*}
\centering
\resizebox{\linewidth}{!}{%
\begin{tabular}{@{}ccccccc@{}}
\toprule
\textbf{Dataset} & \textbf{\# of labels} & \textbf{Input size} & \textbf{\# of Training data} & \textbf{Model} & \textbf{\# Model's output} & \textbf{\# of Parameters total / trainable} \\ \midrule
\multirow{3}{*}{MNIST} & \multirow{3}{*}{10} & \multirow{3}{*}{28 x 28 x 1} & \multirow{3}{*}{60,000} & LeNet prior & 10 & 61706 / 61706 \\
 &  &  &  & LeNet normal & 2 (\textit{even/odd}) & 61026 / 61026 \\
 &  &  &  & LeNet PKT & 2 (\textit{even/odd}) & 61026 / 61026 \\ \midrule
\multirow{3}{*}{CIFAR-10} & \multirow{3}{*}{10} & \multirow{3}{*}{32 x 32 x 3} & \multirow{3}{*}{50,000} & VGG11 prior& 10 & 2976938 / 2976938 \\
 &  &  &  & VGG11 normal & 2 (\textit{animal/object}) & 2974882 / 2974882 \\
 &  &  &  & VGG11 PKT & 2 (\textit{animal/object}) &  2974882 / 1772290 \\ \bottomrule
\end{tabular}}
\caption{Information of dataset and models used in our experiments (Sect.~\ref{sec:experiments}). All models trained on MNIST and CIFAR-10 achieve at least 98\% and 91\% test-set accuracy, respectively. The number of trainable parameters in VGG11 PKT is not all the parameters due to prior-knowledge training.}
\label{tab:dataset-model}
\end{table*}

\begin{table*}[]
\centering

\begin{tabular}{@{}llccccc@{}}
\toprule
Model & Layer & NeuCEPT & Saliency & IntegratedGradients & DeepLift & GradientSHAP \\ \midrule
\multirow{2}{*}{LeNet prior} & conv3 & 3.18 & 17.35 & 79.59 & 51.32 & 31.42 \\
 & linear0 & 1.93 & 16.25 & 79.20 & 20.12 & 31.13 \\
\multirow{2}{*}{LeNet normal} & conv3 & 2.86 & 16.52 & 79.95 & 51.22 & 31.03 \\
 & linear0 & 2.00 & 15.47 & 77.57 & 19.93 & 31.08 \\
\multirow{2}{*}{LeNet PKT} & conv3 & 2.76 & 16.00 & 80.46 & 50.49 & 31.93 \\
 & linear0 & 1.94 & 15.53 & 77.13 & 20.05 & 30.77 \\ \midrule
\multirow{2}{*}{VGG11 prior} & conv8 & 20.86 & 167.82 & 379.16 & 195.01 & 355.65 \\
 & conv9 & 22.36 & 180.6 & 362.28 & 183.97 & 305.65 \\
\multirow{2}{*}{VGG11 normal} & conv8 & 26.12 & 307.74 & 709.38 & 335.29 & 708.50 \\
 & conv9 & 23.22 & 307.87 & 708.71 & 309.38 & 707.38 \\
\multirow{2}{*}{VGG11 PKT} & conv8 & 17.26 & 223.24 & 457.78 & 263.63 & 457.09 \\
 & conv9 & 18.80 & 222.50 & 457.81 & 220.53 & 457.32 \\ \bottomrule
\end{tabular}
\caption{The running-time (in seconds) of tested methods. For the 4 other explanation methods, we use the Captum library which runs on GPUs. Our NeuCEPT runs entirely on CPU cores and the reported running-time is for one iteration. For more stable results, the experimental results of NeuCEPT are obtained after 50 iterations.}
\label{tab:runtime}
\end{table*}

\begin{figure*}
    \centering
    \includegraphics[width=0.99\linewidth]{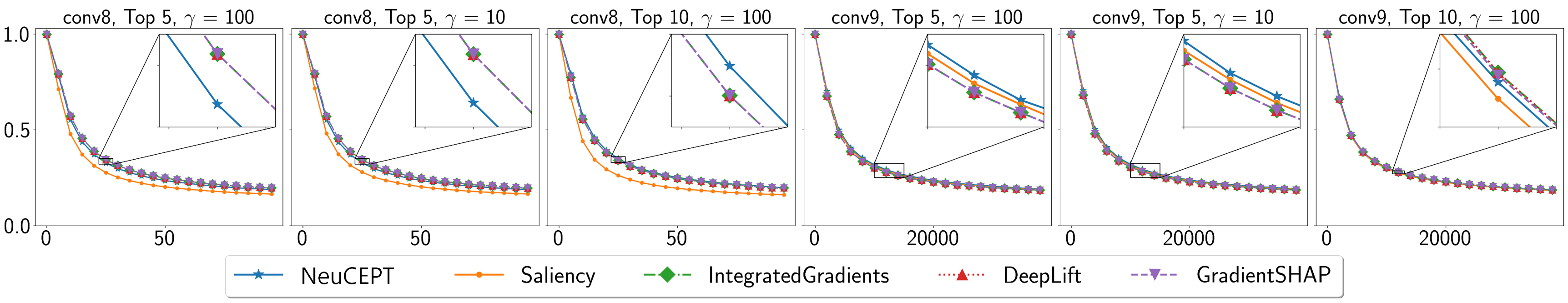}
    \caption{Ablation test of VGG. The x-axis and y-axis are the noise levels and the test accuracy, respectively.}
    \label{fig:ablation}
\end{figure*}

\begin{figure*}
    \centering
    \includegraphics[width=0.99\linewidth]{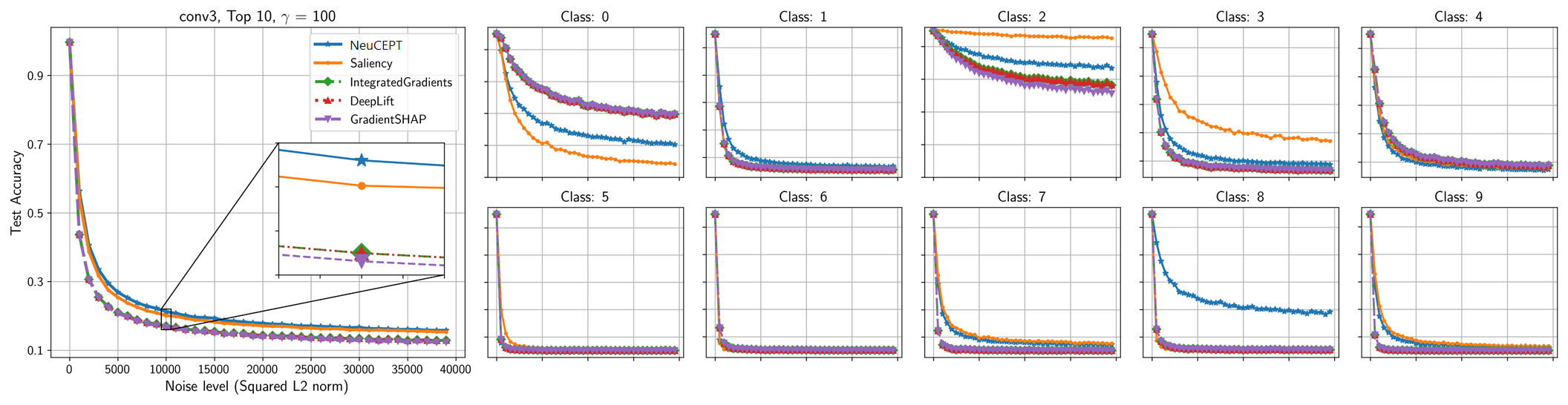}
    \caption{Average and all classes' ablation results at layer \textit{conv3} of LeNet.}
    \label{fig:ablation_lenet_conv3}
\end{figure*}

\begin{figure*}
    \centering
    \includegraphics[width=0.99\linewidth]{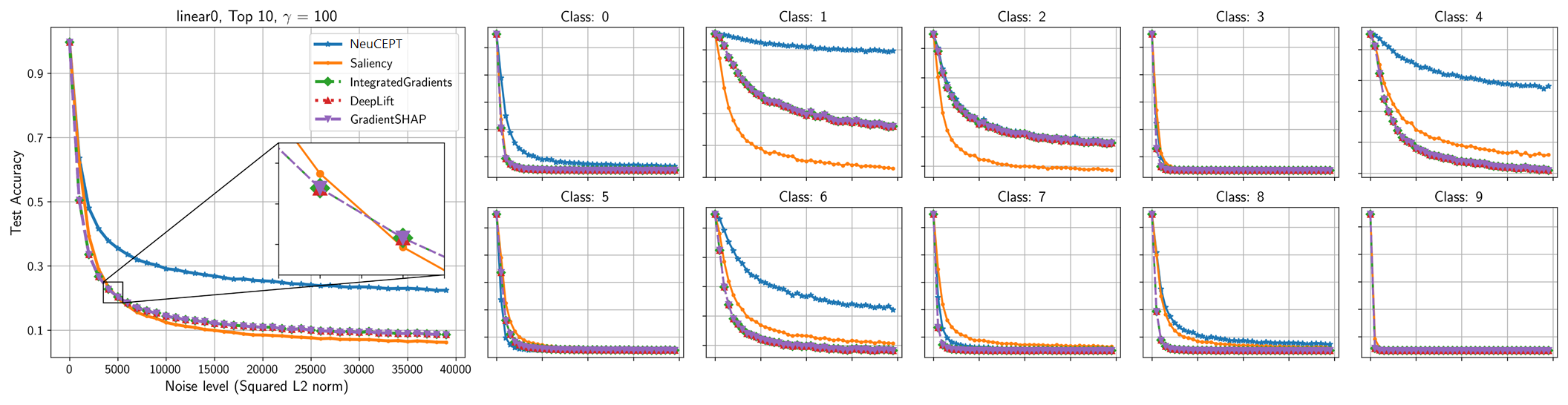}
    \caption{Average and all classes' ablation results at layer \textit{linear0} of LeNet.}
    \label{fig:ablation_lenet_linear0}
\end{figure*}

\begin{figure*}
    \centering
    \includegraphics[width=0.99\linewidth]{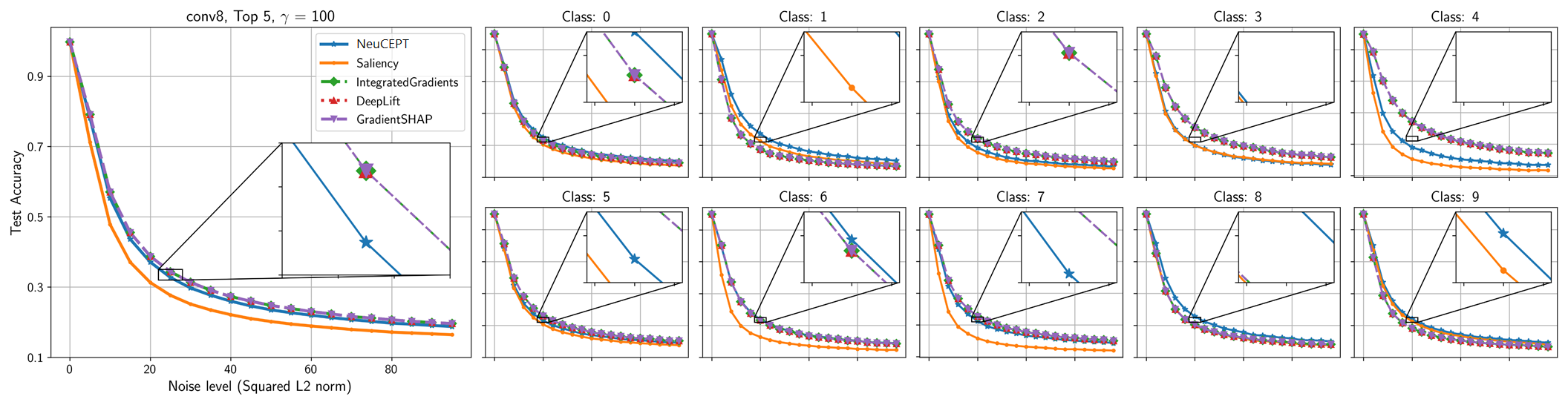}
    \caption{Average and all classes' ablation results at layer \textit{conv8} of VGG.}
    \label{fig:ablation_vgg_conv8}
\end{figure*}

\begin{figure*}
    \centering
    \includegraphics[width=0.99\linewidth]{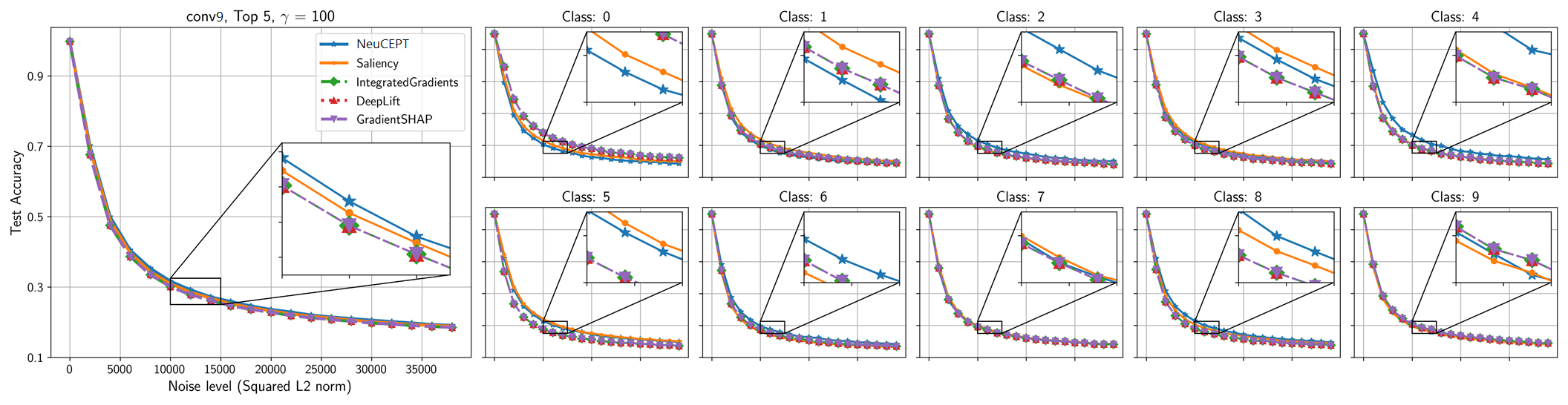}
    \caption{Average and all classes' ablation results at layer \textit{conv9} of VGG.}
    \label{fig:ablation_vgg_conv9}
\end{figure*}

\begin{figure*}
    \centering
    \includegraphics[width=0.99\linewidth]{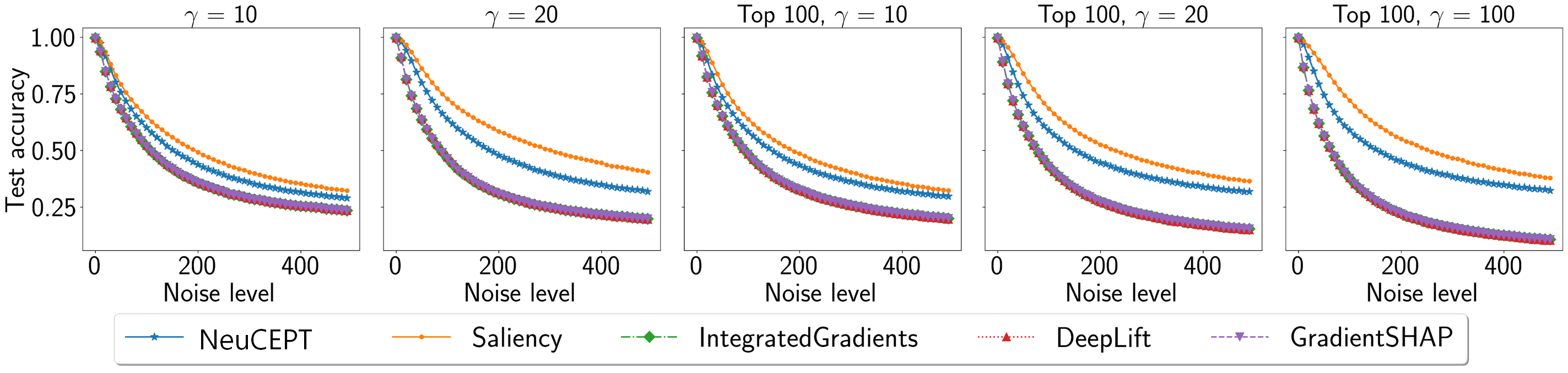}
    \caption{Ablation results at input layer of LeNet. The two plots on the left show the ablation results with \textit{continuous} noise, i.e. the noise are added based on the score given by the associated methods. The three figures on the right shows results when top 100 features are selected. $\gamma$ is an exponential decay parameter determining how noise is distributed among neurons.}
    \label{fig:ablation_lenet_input}
\end{figure*}

\section{Experiments: settings and implementations}\label{appendix:experiment}
Our experiments in this paper are implemented using Python 3.8 and conducted on a single GPU-assisted compute node that is installed with a Linux 64-bit operating system. The allocated resources include 32 CPU cores (AMD EPYC 7742 model) with 2 threads per core, and 100GB of RAM. The node is also equipped with 8 GPUs (NVIDIA DGX A100 SuperPod model), with 80GB of memory per GPU.

\subsection{Model and dataset} 
We experiment with LeNet~\cite{Lecun1998} and the VGG~\cite{Liu2015VeryDC} trained on the MNIST~\cite{lecun2010} and CIFAR10~\cite{Krizhevsky2009} dataset, respectively. The details of datasets and model architectures are listed in Table \ref{tab:dataset-model}. For LeNet, we use the default architecture provided by Pytorch~\cite{pytorch} with only the modification at the last layer. Specifically, we change the last layer to 2 neurons so that the model can be suitable to the even/odd classification posterior-task. In CIFAR-10, we implement the VGG11s using Pytorch along with the modification at the last layer to fit to the \textit{animal/object} classification task, i.e. set the number of neurons at the last layer to 2. 

Our experiments on Inception-v3~\cite{Inception2015} and CheXNet~\cite{CheXNet} are for visualization and demonstration of NeuCEPT on a wider range of models and dataset. We use the pre-trained Inception-v3 provided by Pytorch and test the model on the Imagenet dataset~\cite{imagenet_cvpr09}. The CheXNet is provided by the original paper~\cite{CheXNet}.


\subsection{Implementation details}

 \textbf{Freezing parameters during prior-knowledge training.} As indicated in Table.~\ref{tab:dataset-model}, all LeNets' parameters during prior-knowledge training on the posterior task are trainable. However, in the training of the PKT model for the posterior-task of CIFAR10, we freeze the PKT model's parameters up to the $6^{\textup{th}}$ convolutional layer (there are 9 convolutional layers), i.e., only about $60\%$ of the model parameters are trainable in this task. The reason for this is we want the prior-knowledge obtained from the weights' transferring (Fig. 6) remains in the PKT model.

\textbf{Baseline.} To our knowledge, there exists no work directly addressing the proposed problem. As such, we adopt state-of-the-art local explanation methods, including Saliency~\cite{Simonyan2013}, IntegratedGradients~\cite{Mukund2017}, Deeplift~\cite{Avanti2017} and GradientSHAP~\cite{Scott2017}, implemented by Captum~\cite{Kokhlikyan2020CaptumAU}, to identify critical neurons and evaluate the corresponding identified mechanisms accordingly. As they are back-propagation methods, the neurons' attribution on each input can be extracted in a straight-forward manner, i.e. we can simply back-propagate to the examined layer. The neurons are then selected by each method based on their total attribution scores on all images of the examined class.

\textbf{The choice of precision values.} There are several factors deciding the choices of precision threshold in the identification of critical neurons. The main three factors are the layer's position, the number of neurons in the layer and the overall complexity of the model. The intuition is that, the latter the layer, the smaller of number of neurons in the layer and the less complexity the model, the easier we can identify critical neurons. Thus, in such case, we can set a lower value of FDR and expect better solution. In experiments for LeNet, the precision is between $0.9$ and $0.98$. For CIFAR10, we select values between $0.4$ and $0.8$ based on the layer and the number of neurons. In Inception-v3 and CheXNet, the values are $0.6$ for all layers.


\subsection{Ablation test} \label{appendix:ablation}

Since the mechanism specifies the prediction, critical neurons should have high impact on model's predictions. This impact can be evaluated via ablation test, in which noise of different levels and configurations are added to those neurons and the model's accuracy is recorded accordingly. If the neurons have explainability power, protecting them from the noise should maintain the model's predictive power.

\textbf{Noise generation.} In ablation test, we first generate an uniformly random base noise $\delta_i \in [0,1]$ for each neuron $i$ of testing. Then, the weighted noise for each neuron is computed as $ \delta_i 2^{-\gamma s_i}$, where $\gamma$ is a parameter defined the noise and $s_i$ is the normalized importance score of the neuron. The actual values of the scores are determined by the explanation methods and how the neurons are selected. For example, if we test the top-k neurons, then the scores for those neurons are 1 and the rest are 0. The final noise added on the neurons is computed by normalizing the weighted noise based on the specified noise level. Note that in each computation of an input, only one $\delta_i$ is generated for all explanation methods; thus, the differences in the actual noises added onto the neurons across methods are introduced only by the difference in the score $s_i$, not by the source of the random generator. 

\textbf{Results.} Fig.~\ref{fig:ablation_small} and~\ref{fig:ablation} shows our ablation tests of neurons identified by different methods. In the experiments, \textit{Top} is the number of neurons protected from the noise, determined by the score of the explanation methods.  $\gamma$ is an exponential decay parameter determining how noise is distributed among neurons: the larger the $\gamma$, the lesser the noise added to the protected neurons. The test is conducted at different layers of LeNet and VGG. The results show that neurons identified by NeuCEPT hold higher predictive power among most of the experiments. A more distinctive difference among methods can be observed at the ablation results for each class label in Fig.~\ref{fig:ablation_lenet_conv3}, \ref{fig:ablation_lenet_linear0}, \ref{fig:ablation_vgg_conv8} and \ref{fig:ablation_vgg_conv9}. The results imply that the methods select neurons differently even though the averaging ablation's results can be similar. 

\textbf{NeuCEPT on input features.} Fig.~\ref{fig:ablation_lenet_input} demonstrates our adaptation of NeuCEPT to LeNet to select important input features for each class. The purpose of this experiment is to demonstrate that NeuCEPT can adapt to the concerns regarding Model-X Knockoffs (Appendix~\ref{subsect:modelx}), which are high-dimensional highly-correlated data and unknown covariance, in the context of neural network. Here, we apply Model-X Knockoffs' \textit{representative} approach: dividing the features into $4 \times 4$ groups based on their spatial positions on the image and using a representative variable for each group. The activation/realization of that variable is the average activation of features in the group. Then, those representative neurons are considered as the inputs for NeuCEPT. Fig.~\ref{fig:ablation_lenet_input} shows the ablation results of neurons selected by NeuCEPT, compared with other methods. The plots show that the explainability power of features selected by NeuCEPT is comparable with other local explanation methods.

\subsection{Running-Time.}
We provide the experimental running-time of NeuCEPT and other methods in Table~\ref{tab:runtime}. Precisely, it is the time to obtain the set of important neurons for the class of examination. All the experiments are conducted on 10000 samples of test set. A straight comparison among methods is not trivial due to the difference in resource utilization: our method runs entirely on CPUs while some other methods utilize both CPUs and GPUs. Hyper-parameters selections also affect the running-time of some methods significantly. Technically, in NeuCEPT, we only need to run Model-X Knockoffs once and obtain the critical neurons; however, we run Model-X Knockoffs 50 times in our experiments for more stable results. The main takeaway is that NeuCEPT can be run in a reasonable amount of time on moderate-size models such as VGG.

\end{document}